\documentclass[nohyperref]{article}

\usepackage{microtype}
\usepackage{graphicx}
\usepackage{booktabs} %
\usepackage{adjustbox}
\usepackage[external,edges]{forest}

\usepackage{tikz-qtree}
\usetikzlibrary{arrows.meta}
\usepackage{framed}
\usepackage{enumitem}
\usepackage{tablefootnote}
\usepackage{listings}
\usepackage[most]{tcolorbox}
\usepackage{lmodern}
\usepackage[T1]{fontenc}
\usepackage[framemethod=TikZ]{mdframed}
\newtcbox{\mybox}[1][red]{on line,
    colback=#1, colframe=#1, boxsep=0pt, boxrule=0pt, size=small, arc=1mm}

\newcommand*\rot{\rotatebox{50}}

\usepackage{pgfplotstable,filecontents}
\pgfplotstableset{fixed zerofill,precision=2}
\usepackage{url}
\usepackage{times}
\usepackage{graphicx}
\usepackage{appendix}
\usepackage{pgfplotstable,filecontents}
\usepackage{booktabs}
\usepackage{caption}
\usepackage{subcaption}
\usepackage{adjustbox}
\usepackage{float}
\pgfplotstableset{fixed zerofill,precision=2}
\usepackage[section]{placeins}
\pgfplotsset{compat=1.9}%

\usepackage{hyperref}

\usepackage[accepted]{icml2022}

\usepackage{amsmath}
\usepackage{amssymb}
\usepackage{mathtools}
\usepackage{amsthm}
\usepackage{array}
\usepackage{ragged2e}
\newcolumntype{P}[1]{>{\RaggedRight\hspace{0pt}}m{#1}}
\usepackage{color, colortbl}
\definecolor{Gray}{gray}{0.94}

\usepackage[capitalize,noabbrev]{cleveref}
\theoremstyle{plain}

\theoremstyle{definition}

\theoremstyle{remark}

\usepackage[textsize=tiny]{todonotes}

\icmltitlerunning{Leakage and the Reproducibility Crisis in ML-based Science}
\icmlauthorrunning{Kapoor and Narayanan}

\begin{document}

\twocolumn[
\icmltitle{Leakage and the Reproducibility Crisis in ML-based Science}

\begin{icmlauthorlist}
\icmlauthor{Sayash Kapoor}{pton}
\icmlauthor{Arvind Narayanan}{pton}
\end{icmlauthorlist}

\icmlaffiliation{pton}{Department of Computer Science and Center for Information Technology Policy, Princeton University}

\icmlcorrespondingauthor{Sayash Kapoor}{sayashk@princeton.edu}

\icmlkeywords{ML-based science, reproducibility, data leakage, model info sheets}

\vskip 0.3in
]

\printAffiliationsAndNotice{}  %

\begin{abstract}
    The use of machine learning (ML) methods for prediction and forecasting has become widespread across the quantitative sciences. However, there are many known methodological pitfalls, including data leakage, in ML-based science. In this paper, we systematically investigate reproducibility issues in ML-based science. We show that data leakage is indeed a widespread problem and has led to severe reproducibility failures. Specifically, through a survey of literature in research communities that adopted ML methods, we find 17 fields where errors have been found, collectively affecting 329 papers and in some cases leading to wildly overoptimistic conclusions. Based on our survey, we present a fine-grained taxonomy of 8 types of leakage that range from textbook errors to open research problems. 
    
    We argue for fundamental methodological changes to ML-based science so that cases of leakage can be caught before publication. To that end, we propose model info sheets for reporting scientific claims based on ML models that would address all types of leakage identified in our survey. To investigate the impact of reproducibility errors and the efficacy of model info sheets, we undertake a reproducibility study in a field where complex ML models are believed to vastly outperform older statistical models such as Logistic Regression (LR): civil war prediction. We find that all papers claiming the superior performance of complex ML models compared to LR models fail to reproduce due to data leakage, and complex ML models don't perform substantively better than decades-old LR models. While none of these errors could have been caught by reading the papers, model info sheets would enable the detection of leakage in each case. 
\end{abstract}

\section{Overview}
\label{intro}

There has been a marked shift towards the paradigm of predictive modeling across quantitative science fields. 
This shift has been facilitated by the widespread use of machine learning (ML) methods.
However, pitfalls in using ML methods have led to exaggerated claims about their performance. Such errors can lead to a feedback loop of overoptimism about the paradigm of prediction---especially as non-replicable publications tend to be cited more often than replicable ones \cite{serra-garcia_nonreplicable_2021}.
It is therefore important to examine the reproducibility of findings in communities adopting ML methods.

\begin{table*}[t]
  \centering
  \footnotesize
  \begin{adjustbox}{width=\textwidth,center}
\begin{tabular}{p{2.5cm}p{3.4cm}p{0.2cm}p{0.2cm}p{0.2cm}p{0.2cm}p{0.2cm}p{0.2cm}p{0.2cm}p{0.2cm}p{0.2cm}p{0.2cm}p{0.2cm}p{0.2cm}p{0.2cm}p{0.5cm}}
{Field}&{Paper}&\rot{Number of papers reviewed}&\rot{Number of papers with pitfalls}&\rot{\colorbox{white}{\textbf{[L1.1] No test set}}}&\rot{\colorbox{white}{\textbf{[L1.2] Pre-proc. on train-test}}}&\rot{\colorbox{white}{\textbf{[L1.3] Feature sel. on train-test}}}&\rot{\colorbox{white}{\textbf{[L1.4] Duplicates}}}&\rot{\colorbox{white}{\textbf{[L2] Illegitimate features}}}&\rot{\colorbox{white}{\textbf{[L3.1] Temporal leakage}}}&\rot{\colorbox{white}{\textbf{[L3.2] Non-ind. b/w train-test}}}&\rot{\colorbox{white}{\textbf{[L3.3] Sampling bias}}}&\rot{Comput. reproducibility issues}&\rot{Data quality issues}&\rot{Metric choice issues}&\rot{Standard dataset used?}
\\ 
\midrule
Medicine & \citet{bouwmeester_reporting_2012}&71&27&$\circ$&&&&&&&&&$\circ$&&
\\ \rowcolor{Gray}
Neuroimaging& \citet{whelan_when_2014}&--&14&$\circ$&&$\circ$&&&&&&&&&  
\\
Autism Diagnostics& \citet{bone_applying_2015}&--&3&&&&$\circ$&&&&$\circ$&&$\circ$&$\circ$&$\circ$ 
\\ \rowcolor{Gray}
Bioinformatics& \citet{blagus_joint_2015}&--&6&&$\circ$&&&&&&&&&& 
\\
Nutrition Research& \citet{ivanescu_importance_2016}&--&4&$\circ$&&&&&&&&&$\circ$&$\circ$& 
\\ \rowcolor{Gray}
Software Eng.& \citet{tu_be_2018}&58&11&&&&&&$\circ$&&&$\circ$&$\circ$&&$\circ$ 
\\
Toxicology&\citet{alves_oy_2019}&--&1&&&&$\circ$&&&&&$\circ$&$\circ$&& 
\\ \rowcolor{Gray}
Satellite Imaging&\citet{nalepa_validating_2019}&17&17&&&&&&&$\circ$&&&$\circ$&&$\circ$ 
\\
Tractography& \citet{poulin_tractography_2019}&4&2&$\circ$&&&&&&&&$\circ$&$\circ$&$\circ$&$\circ$
\\ \rowcolor{Gray}
Clinical Epidem.& \citet{christodoulou_systematic_2019}&71&48&&&$\circ$&&&&&&&$\circ$&& 
\\ 
Brain-computer Int.& \citet{nakanishi_questionable_2020}&--&1&$\circ$&&&&&&&&&&&$\circ$ 
\\\rowcolor{Gray}
Histopathology& \citet{oner_training_2020}&--&1&&&&&&&$\circ$&&&&& 
\\ 
Neuropsychiatry& \citet{poldrack_establishment_2020}&100&53&$\circ$&$\circ$&&&&&&&&$\circ$&$\circ$& 
\\ \rowcolor{Gray}
Medicine& \citet{vandewiele_overly_2021}&24&21&&&$\circ$&&&&$\circ$&$\circ$&$\circ$&$\circ$&&$\circ$ 
\\
Radiology& \citet{roberts_common_2021}&62&62&$\circ$&&&$\circ$&&&&$\circ$&$\circ$&&&$\circ$ 
\\ \rowcolor{Gray}
IT Operations& \citet{lyu_empirical_2021}&9&3&&&&&&$\circ$&&&&&&$\circ$ 
\\
Medicine& \citet{filho_data_2021}&--&1&&&&&$\circ$&&&&&&&
\\ \rowcolor{Gray}
Neuropsychiatry& \citet{shim_inflated_2021}&--&1&&&$\circ$&&&&&&$\circ$&&& 
\\ 
Genomics& \citet{barnett_genomic_2022}&41&23&&&$\circ$&&&&&&&$\circ$&& 
\\ \rowcolor{Gray}
Computer Security& \citet{arp_dos_2022}&30&30&$\circ$&$\circ$&$\circ$&&$\circ$&&$\circ$&$\circ$&&$\circ$&$\circ$& 
\\
\end{tabular}
\end{adjustbox}
\caption{\footnotesize Survey of 20 papers that identify pitfalls in the adoption of ML methods across 17 fields, collectively affecting 329 papers. In each field, papers adopting ML methods suffer from data leakage. The column headings for types of data leakage, shown in bold, are based on our taxonomy of data leakage. We also highlight other issues that are reported in the papers, including issues with computational reproducibility (the availability of code, data, and computing environment to reproduce the exact results reported in the paper), data quality (for example, small size or large amounts of missing data), metric choice (using incorrect metrics for the task at hand, for example, using accuracy for measuring model performance in the presence of heavy class imbalance), and standard dataset use, where issues are found despite the use of standard datasets in a field. }
  \label{table:survey}
  \vspace{-0.3cm}
\end{table*}

\begin{figure*}[t]
    \centering
    \includegraphics[width=\textwidth]{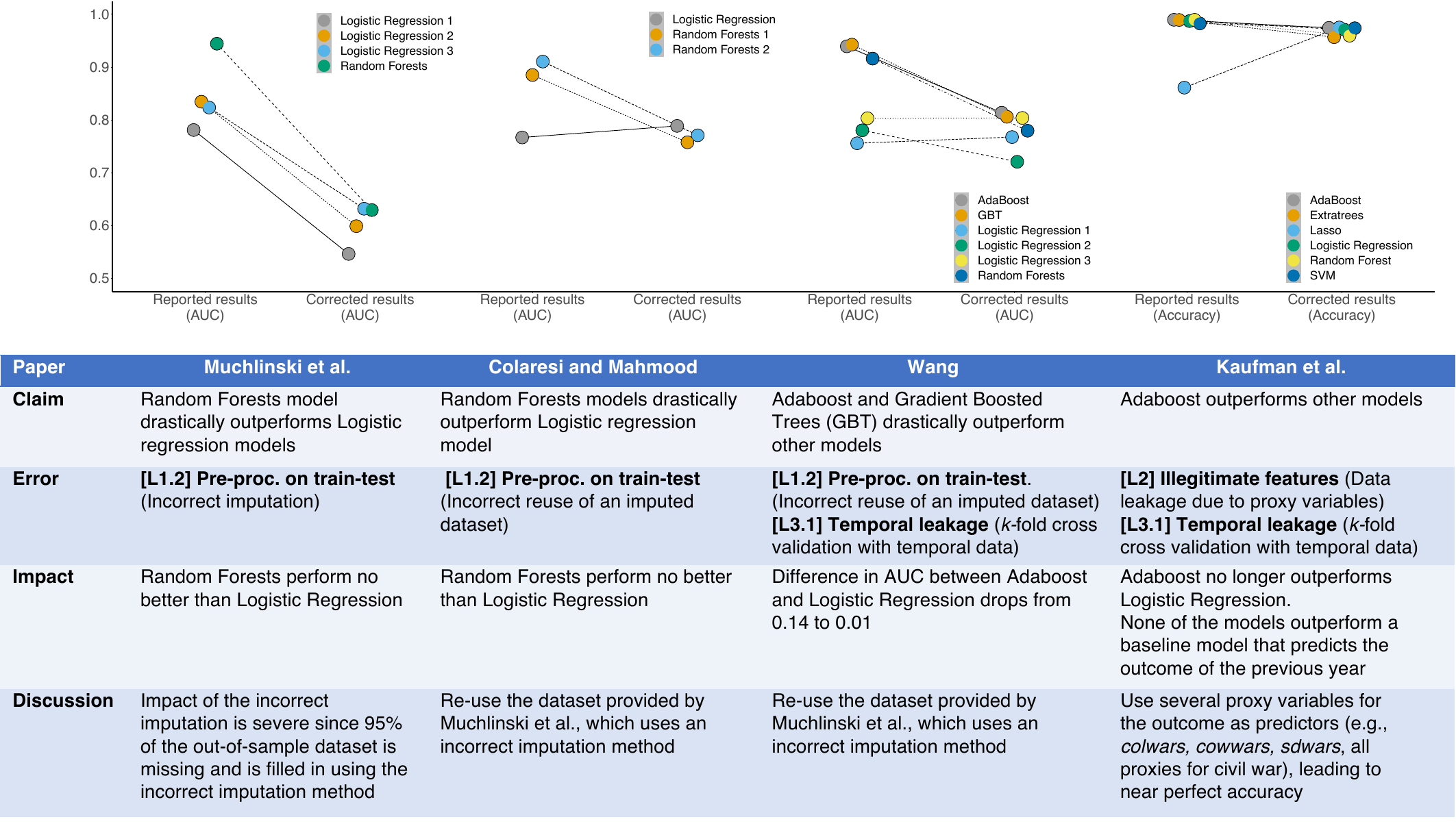}
    \vspace{-1cm}
    \caption{
        A comparison of reported and corrected results in civil war prediction papers published in top political science journals.
        The main findings of each of these papers are invalid due to various forms of data leakage: 
        \citet{muchlinski_comparing_2016} impute the training and test data together, 
        \citet{colaresi_robot_2017} and \citet{wang_comparing_2019} incorrectly reuse an imputed dataset, 
        and \citet{kaufman_improving_2019} use proxies for the target variable which causes data leakage. 
        The use of model info sheets (\cref{sec:model_cards}) would detect leakage in every paper.
        When we correct these errors, complex ML models (such as Adaboost and Random Forests) do not perform substantively better than decades-old Logistic Regression models for civil war prediction in each case.
        Each column in the table outlines the impact of leakage on the results of a paper. The figure above each column shows the difference in performance that results from fixing leakage issues.
    }
    \label{fig:allFive}
\end{figure*}

{\bf Scope.} We focus on reproducibility issues in ML-based science, which involves making a scientific claim using the performance of the ML model as evidence. 
There is a much better known reproducibility crisis in research that uses traditional statistical methods \cite{open_science_collaboration_estimating_2015}.
We also situate our work in contrast to other ML domains, such as methods research (creating and improving widely-applicable ML methods), ethics research (studying the ethical implications of ML methods), engineering applications (building or improving a product or service), and modeling contests (improving predictive performance on a fixed dataset created by an independent third party). Investigating the validity of claims in all of these areas is important, and there is ongoing work to address reproducibility issues in these domains \cite{hullman_worst_2022, pineau_improving_2020, erik_gundersen_fundamental_2021, bell_perspectives_2021}. 

We define a research finding as reproducible if the code and data used to obtain the finding are available and the data is correctly analyzed \cite{hofman_expanding_2021, leek_opinion_2015, pineau_improving_2020}. This is a broader definition than computational reproducibility --- when the results in a paper can be replicated using the exact code and dataset provided by the authors (see Appendix \ref{app:terminology}).

{\bf Leakage.} Data leakage has long been recognized as a leading cause of errors in ML applications \cite{nisbet_handbook_2009}. In formative work on leakage, \citet{kaufman_leakage_2012} provide an overview of different types of errors and give several recommendations for mitigating these errors. Since this paper was published, the ML community has investigated leakage in several engineering applications and modeling competitions \cite{fraser_treachery_2016, ghani_top_2020, becker_data_2018, brownlee_data_2016, collins-thompson_data_nodate}. However, leakage occurring in ML-based science has not been comprehensively investigated. As a result, mitigations for data leakage in scientific applications of ML remain understudied. 

In this paper, we systematically investigate reproducibility issues in ML-based science due to data leakage. We make three main contributions:

\textbf{1) A survey and taxonomy of reproducibility issues due to leakage.}
We provide evidence for a growing reproducibility crisis in ML-based science. Through a survey of literature in research communities that adopted ML methods, we find 20 papers across 17 fields where errors have been found, collectively affecting 329 papers (\cref{table:survey}). Each of these fields suffers from leakage. 
We highlight that data leakage mitigation strategies developed for other ML applications such as modeling contests and engineering applications often do not translate to ML-based science. 
Based on our survey, we present a fine-grained taxonomy of 8 types of leakage that range from textbook errors to open research problems (\cref{sec:taxonomy}). 

\textbf{2) Model info sheets to detect and prevent leakage.}
Current standards for reporting model performance in ML-based science often fall short in addressing issues due to leakage. 
Specifically, checklists and model cards are one way to provide standard best practices for reporting details about ML models \cite{mongan_checklist_2020, collins_transparent_2015, mitchell_model_2019}. 
However, current efforts do not address issues arising due to leakage. Further, most checklists currently in use are not developed for ML-based science in general, but rather for specific scientific or research communities \cite{pineau_improving_2020, mongan_checklist_2020}. As a result, best practices for model reporting in ML-based science are underspecified.

In this paper, we introduce model info sheets to detect and prevent leakage in ML-based science (\cref{sec:model_cards}). 
They are inspired by the model cards in \citet{mitchell_model_2019}. Filling out a model info sheet requires the researcher to provide precise arguments to justify that models used towards making scientific claims do not suffer from leakage. 
Model info sheets address all types of leakage identified in our survey.
We advocate for model info sheets to be included with every paper making a scientific claim using an ML model.

\textbf{3) Empirical case study of leakage in civil war prediction.}
For an in-depth look at the impact of reproducibility errors and the efficacy of model info sheets, we undertake a reproducibility study in civil war prediction, a subfield of political science where ML models are believed to vastly outperform older statistical models such as Logistic Regression. 
We perform a systematic review to find papers on civil war prediction and find that all papers in our review claiming the superior performance of ML models compared to Logistic Regression models fail to reproduce due to data leakage (\cref{fig:allFive})\footnote{\textbf{A note on terminology.} We use ``ML models'' as a shorthand for models other than Logistic Regression, specifically, Random Forests, Gradient-Boosted Trees, and Adaboost. To be clear, all of these models including Logistic Regression involve learning from data in the predictive modeling approach. However, the terminology we use is common in fields that distinguish ML from statistical methods that invoke an assumption about the true data generating process, such as Logistic Regression  \cite{christodoulou_systematic_2019}.}.
Each of these papers was published in top political science journals. Further, when the errors are corrected, ML models don’t perform substantively better than decades-old Logistic Regression models, calling into question the shift from explanatory modeling to predictive modeling in this field. While none of these errors could have been caught by reading the papers, model info sheets enable the detection of leakage in each case.

\section{Evidence of a reproducibility crisis}
Many scientific fields have adopted ML methods and the paradigm of predictive modeling  \cite{athey_machine_2019,schrider_supervised_2018,valletta_applications_2017,iniesta_machine_2016,tonidandel_big_2018, yarkoni_choosing_2017}.
We find at least three main uses of ML models in scientific literature. 
First, models which are better at prediction are thought to enable an improved understanding of scientific phenomena \cite{hofman_integrating_2021}.
Second, especially when used in medical fields, models with higher predictive accuracy can aid in research and development of better diagnostic tools \cite{mcdermott_reproducibility_2021}.
Finally, ML-based methods have also been used to investigate the inherent predictability of phenomena, especially for predicting social outcomes \cite{salganik_measuring_2020}. The increased adoption of ML methods in science motivates our investigation of reproducibility issues in ML-based science.

\subsection{Data leakage causes irreproducible results} \label{subsec:leakage_irrep}

Data leakage is a spurious relationship between the independent variables and the target variable that arises as an artifact of the data collection, sampling, or pre-processing strategy. Since the spurious relationship won’t be present in the distribution about which scientific claims are made, leakage usually leads to inflated estimates of model performance. 

Researchers in many communities have already documented reproducibility failures in ML-based science within their fields. Here we conduct a cross-disciplinary analysis by building on these individual reviews. This enables us to highlight the scale and scope of the crisis, identify common patterns, and make progress toward a solution. 

When searching for past literature that documents reproducibility failures in ML-based science, we found that different fields often use different terms to describe pitfalls and errors. This makes it difficult to conduct a systematic search to find  papers with errors.
Therefore, we do not present our results as a systematic meta-review of leakage from a coherent sample of papers, but rather as a lower bound of reproducibility issues in ML-based science.
Additionally, most reviews only look at the content of the papers, and not the code and data provided with the papers to check for errors. This leads to under-counting the number of affected papers, since the code might have errors that are not apparent from reading the papers. 

Our findings present a worrying trend for the reproducibility of ML-based science. We find 20 papers from 17 fields that outline errors in ML-based science in their field, collectively affecting 329 papers. A prominent finding that emerges is that data leakage is a pitfall in every single case. The results from our survey are presented in  \cref{table:survey}. Columns in bold represent different types of leakage (\cref{sec:taxonomy}). The last four columns represent other common trends in the papers we study (\cref{subsec:other_issues}). For systematic reviews, we report the number of papers reviewed. Each paper in our survey highlights issues with leakage, with 6 papers highlighting the presence of multiple types of leakage in their field.

\subsection{Data leakage mitigations for other ML applications do not apply to scientific research}
\label{sec:leakage_understudied}

Most previous research and writing on data leakage has focused on mitigating data leakage primarily for engineering settings or predictive modeling competitions \cite{kaufman_leakage_2012}. 
However, the taxonomy of data leakage outlined in this body of work does not address all kinds of leakage that we identify in our survey. 
In particular, we find that leakage can result from a difference between the distribution of the test set and the distribution of scientific interest (\cref{sec:taxonomy}). Robustness to distribution shift is an area of ongoing research in ML methods, and is as such an open problem \cite{geirhos_shortcut_2020}.
Additionally, prior work primarily focuses on mitigating leakage in modeling competitions and engineering applications.
Both of these settings are very different from scientific research, and mitigations for data leakage in modeling competitions as well as engineering applications of ML often do not translate into strategies for mitigating data leakage in ML-based science. 

\noindent\textbf{Leakage in modeling competitions.} In predictive modeling competitions, dataset creation and model evaluation is left to impartial third parties who have the expertise and incentives to avoid errors. Within this framework, none of the participants have access to the held-out evaluation set before the competition ends. In contrast, in most ML-based science the researcher has access to the entire dataset while creating the ML models. Leakage often occurs due to the researcher having access to the entire dataset during the modeling process.

\noindent\textbf{Leakage in engineering applications.} One of the most common recommendations for detecting and mitigating leakage is to deploy the ML model at a limited scale in production. This advice is only applicable to engineering applications of ML, where the end goal is not to gain insights about a particular process, but rather to serve as a component in a product. Often, a rough idea of model performance is enough to decide whether a model is good enough to be deployed in a product. Contrarily, ML-based science involves making a scientific claim using the performance of the ML model as evidence. In addition, engineering applications of ML often operate in a rapidly changing context and have access to large datasets, so small differences in performances are often not as important, whereas scientific claims are sensitive to small performance differences between ML models.

\subsection{Why do we call it a reproducibility crisis?}
We say that ML-based science is suffering from a reproducibility crisis for two related reasons:
First, our results show that reproducibility failures in ML-based science are systemic. In nearly every scientific field that has carried out a systematic study of reproducibility issues, papers are plagued by common pitfalls. 
In many systematic reviews, a majority of the papers reviewed suffer from these pitfalls. Thus, we find that similar problems are likely to arise in many fields that are adopting ML methods.
Second, despite the urgency of addressing reproducibility failures, there are no systemic solutions that have been deployed for these failures. 
Scientific communities are discovering the same failure modes across disciplines, but have yet to converge on best practices for avoiding reproducibility failures. 

Calling attention to and addressing these widespread failures is vital to maintaining public confidence in ML-based science.
At the same time, the use of ML methods is still in its infancy in many scientific fields. Addressing reproducibility failures pre-emptively in such fields can correct a lot of scientific research that would otherwise be flawed.

\subsection{Towards a solution: A taxonomy of data leakage} \label{sec:taxonomy}

We now provide our taxonomy of data leakage errors in ML-based science. 
Such a taxonomy can enable a better understanding of why leakage occurs and inform potential solutions. Our taxonomy is comprehensive and addresses data leakage arising during the data collection, pre-processing, modeling and evaluation steps. In particular, our taxonomy addresses all cases of data leakage that we found in our survey~(\cref{table:survey}). 

\textbf{[L1] Lack of clean separation of training and test dataset.} 
If the training dataset is not separated from the test dataset during all pre-processing, modeling and evaluation steps, the model has access to information in the test set before its performance is evaluated. Since the model has access to information from the test set at training time, the model learns relationships between the predictors and the outcome that would not be available in additional data drawn from the distribution of interest. The performance of the model on this data therefore does not reflect how well the model would perform on a new test set drawn from the same distribution of data.
 
\textit{[L1.1] No test set.} Using the same dataset for training as well as testing the model is a text-book example of overfitting, which leads to overoptimisic performance estimates \cite{kuhn_applied_2013}. 

\textit{[L1.2] Pre-processing on training and test set.} Using the entire dataset for any pre-processing steps such as imputation or over/under sampling. For instance, using oversampling before splitting the data into training and test sets leads to an imperfect separation between the training and test sets since data generated using oversampling from the training set will also be present in the test set.
 
\textit{[L1.3] Feature selection on training and test set.} Feature selection on the entire dataset results in using information about which feature performs well on the test set to make a decision about which features should be included in the model.

\textit{[L1.4] Duplicates in datasets.} If a dataset with duplicates is used for the purposes of training and evaluating an ML model, the same data could exist in the training as well as test set. 

\textbf{[L2] Model uses features that are not legitimate.}
If the model has access to features that should not be legitimately available for use in the modeling exercise, this could result in leakage. One instance when this can happen is if a feature is a proxy for the outcome variable \cite{kaufman_leakage_2012}. For example, \citet{filho_data_2021} find that a recent study included the use of anti-hypertensive drugs as a feature for predicting hypertension. Such a feature could lead to leakage because the model would not have access to this information when predicting the health outcome for a new patient. 
Further, if the fact that a patient uses anti-hypertensive drugs is already known at prediction time, the prediction of hypertension becomes a trivial task.

The judgement of whether the use of a given feature is legitimate for a modeling task requires domain knowledge and can be highly problem specific. As a result, we do not provide sub-categories for this sort of leakage. Instead, we suggest that researchers decide which features are suitable for a modeling task and justify their choice using domain expertise.

\textbf{[L3] Test set is not drawn from the distribution of scientific interest.} 
The distribution of data on which the performance of an ML model is evaluated differs from the distribution of data about which the scientific claims are made. The performance of the model on the test set does not correspond to its performance on data drawn from the distribution of scientific interest.

\textit{[L3.1] Temporal leakage.} When an ML model is used to make predictions about a future outcome of interest, the test set should not contain any data from a date before the training set. If the test set contains data from before the training set, the model is built using data ``from the future'' that it should not have access to during training, and can cause leakage.

\textit{[L3.2] Nonindependence between train and test samples.} %
Nonindependence between train and test samples constitutes leakage, unless the scientific claim is about a distribution that has the same dependence structure. In the extreme (but unfortunately common) case, train and test samples come from the same people or units.
For example, \citet{oner_training_2020} find that a recent study on histopathology uses different observations of the same patient in the training and test sets. 
In this case, the scientific claim is being made about the ability to predict gene mutations in new patients; however, it is evaluated on data from old patients (i.e., data from patients in the training set), leading to a mismatch between the test set distribution and the scientific claim. 
The train-test split should account for the dependencies in the data to ensure correct performance evaluation. Methods such as `block cross validation' can partition the dataset strategically so that the performance evaluation does not suffer from data leakage and overoptimism \cite{roberts_cross-validation_2017, valavi_block_2021}. Handling nonindependence between the training and test sets in general---i.e., without any assumptions about independence in the data---is a hard problem, since we might not know the underlying dependency structure of the task in many cases \cite{malik_hierarchy_2020}. 

\textit{[L3.3] Sampling bias in test distribution.} 
Sampling bias in the choice of test dataset can lead to data leakage. 
One example of sampling bias is spatial bias, which refers to choosing the test data from a geographic location but making claims about model performance in other geographic locations as well.
Another example is selection bias, which entails choosing a non-representative subset of the dataset for evaluation.
For example, \citet{bone_applying_2015} highlight that in a study on predicting autism using ML models, excluding the data corresponding to borderline cases of autism leads to leakage since the test set is no longer representative of the general population about which claims are made. In addition, borderline cases of autism are often the most tricky to diagnose, so excluding them the evaluation set is likely to lead to overoptimistic results. 
Cases of leakage due to sampling bias can often be subtle. For example, \citet{zech_variable_2018} find that models for pneumonia prediction trained on images from one hospital do not generalize to images from another hospital due to subtle differences in how images are generated in each hospital.

A model may have leakage when the distribution about which the scientific claim is made does not match the distribution from which the evaluation set is drawn. ML models may also suffer from a related, but distinct limitation: the lack of generalization when we try to apply a result about one population to another similar but distinct population.
Several issues with the generalization of ML models operating under a distribution shift have been highlighted in ML methods research,
such as fragility towards adversarial examples \cite{szegedy_intriguing_2014}, image distortion and texture \cite{geirhos_imagenet-trained_2018}, and overinterpretation \cite{carter_overinterpretation_2021}.
Robustness to distribution shift is an ongoing area of work in ML methods research. Even slight shifts in the target distribution can cause performance estimates to change drastically \cite{recht_imagenet_2019}. Despite ongoing work to create ML methods that are robust to distribution shift, best practices to deal with distribution shift currently include testing the ML models on the data from the distribution we want to make claims about \cite{geirhos_shortcut_2020}.
In ML-based science, where the aim is to create generalizable knowledge, we should take results that claim to generalize to a different population from the one models were evaluated on with caution. 

\subsection{Other issues identified in our survey (Table \ref{table:survey})} \label{subsec:other_issues}

\textbf{Computational reproducibility issues.} Computational reproducibility of a finding refers to sharing the complete code and data needed to reproduce the findings reported in a paper exactly. 
This is important to enable external researchers to reproduce results and verify their correctness.
Five papers in our survey outlined the lack of computational reproducibility in their field.

\textbf{Data quality issues.} 
Access to good quality data is essential for creating ML models \cite{paullada2020data, scheuerman_datasets_2021}. Issues with the quality of the dataset could affect the results of ML-based science. 
10 papers in our survey highlighted data quality issues such as not addressing missing values in the data, the small size of datasets compared to the number of predictors, and the outcome variable being a poor proxy for the phenomenon being studied. 

\textbf{Metric choice issues.} A mismatch between the metric used to evaluate performance and the scientific problem of interest leads to issues with performance claims. For example, using accuracy as the evaluation metric with a heavily imbalanced dataset leads to overoptimistic results, since the model can get a high accuracy score by always predicting the majority class. Four papers in our survey highlighted metric choice issues.

\textbf{Use of standard datasets.} Reproducibility issues arose despite the use of standard, widely-used datasets, often because of the lack of standard modeling and evaluation procedures such as fixing the train-test split and evaluation metric for the dataset. Seven papers in our survey highlighted that issues arose despite the use of standard datasets.

\section{Model info sheets for detecting and preventing leakage} \label{sec:model_cards}

Our taxonomy of data leakage highlights several failure modes which are prevalent in ML-based science. 
To detect cases of leakage, we provide a template for a model info sheet to accompany scientific claims using predictive modeling \footnote{The model info sheet template is available on our website: \url{https://reproducible.cs.princeton.edu}}. The template consists of precise arguments needed to justify the absence of leakage.
Model info sheets would address every type of leakage identified in our survey.

\subsection{Prior work on model cards and reporting standards}

Our proposal is inspired by prior work on model cards and checklists, which we now review.

\citet{mitchell_model_2019} introduced model cards for reporting details about ML models, with a focus on precisely reporting the intended use cases of ML models. They also addressed fairness and transparency concerns: they require that the performance of ML models on different groups of users (e.g., on the basis of race, gender, age) is reported and documented transparently. These model cards complement the datasheets introduced by \citet{gebru_datasheets_2021} to document details about datasets in a standard format.

The use of checklists has also been impactful in improving reporting practices in the few fields that have adopted them \cite{han_checklist_2017}. 
While checklists and model cards provide concrete best practices for reporting standards \cite{mongan_checklist_2020, collins_transparent_2015, mitchell_model_2019, garbin_assessing_2022}, current efforts do not address pitfalls arising due to leakage. 
Further, even though several scientific fields---especially those related to medicine---have adopted checklists to improve reporting standards, most checklists are developed for specific scientific or research communities instead of ML-based science in general. %

\subsection{Scientific arguments to surface and prevent leakage} 

When ML models are used to make scientific claims, it is not enough to simply separate the training and test sets and report performance metrics on the test set. 
Unlike research in ML methods, where a model's performance on a hypothetical task (i.e., one that is not linked to a specific scientific claim) is still of interest to the researcher in some cases \cite{raji_ai_2021}, in ML-based science, claims about a model's performance need to be connected to scientific claims using explicit arguments. The burden of proof for ensuring the correctness of these arguments is on the researcher making the scientific claims \cite{lundberg_what_2021}.

In our models, we ask researchers to present three arguments that are essential for determining that scientific results which use ML methods do not suffer from data leakage. 
Note that most ML-based science papers do not present any of the three arguments, although they sometimes partially address the first argument (clean train-test separation) by reporting out-of-sample prediction performance. 
The arguments below are based on our taxonomy of data leakage issues presented in \cref{sec:taxonomy}, and inform the main sections of the model info sheet.

\textbf{[L1] Clean train-test separation.} The researcher needs to argue why the test set does not interact with training data during any of the preprocessing, modeling or evaluation steps to ensure a clean train-test separation.

\textbf{[L2] Each feature in the model is legitimate.} The researcher needs to argue why each feature used in their model is legitimate, i.e., a claim made using each feature is of scientific interest. Note that some models might use hundreds of features. In such cases, it is even more important to reason about the correctness of the features used, since the incorrect use of a single feature in the model can cause leakage. That said, the same argument for why a feature is legitimate can often apply to a whole set of features. For example, for a study using individuals' location history as a feature vector, the use of the entire vector can be justified together. Note that we do not ask for the researcher to list each feature used in their model. Rather, we ask that the justification provided for the legitimacy of the features used in their model should cover every feature used in their model.

\textbf{[L3] Test set is drawn from the distribution of scientific interest.} 
If the distribution about which the scientific claims are made is different from the one on which the model is tested, then any claims about the performance of an ML model on the evaluation step fall short. 
The researcher needs to justify that the test set is drawn from the distribution of scientific interest and there is no selection or sampling bias in the data collection process.
This step can help clarify the distribution about which scientific claims are being made and detect temporal leakage.

\subsection{Model info sheets and our theory of change}

Model info sheets can influence research practices in two ways: first, researchers who introduce a scientific model alongside a paper can use model info sheets to detect and prevent leakage in their models. These info sheets can be included as supplementary materials with their paper for transparently reporting details about their models. 
In scientific fields where the use of ML methods is not yet widespread, using transparent reporting practices at an early stage could enable easier adoption and more trust in ML methods. This would also help assuage reviewer concerns about reproducibility.

Second, journal submission guidelines could encourage or require authors to fill out model info sheets if a paper does not transparently report how the model was created. In this case, model info sheets can be used to start a conversation between authors and reviewers about the details of the models introduced in a paper. Current peer-review practices often do not require the authors to disclose any code or data during the review process \cite{liu_successes_2019}. 
Even if the code and data are available to reviewers, reproducing results and spotting errors in code is a time consuming process that often cannot be carried out under current peer-review practices. Model info sheets offer a middle ground: they could enable a closer scrutiny of  methods without making the process onerous for reviewers.

\subsection{Limitations of model info sheets}\label{sec:model-card-limitations}

While model info sheets can enable the detection of all types of leakage we identify in our survey, they suffer from limitations owing to the lack of computational reproducibility of results in scientific research, incorrect claims made in model info sheets, and the lack of expertise of authors and reviewers. 

First, the claims made in model info sheets cannot be verified in the absence of computational reproducibility. That is, unless the code, data and computing environment required to reproduce the results in a paper are made available, there is no way to ascertain if model info sheets are filled out correctly. Ensuring the computational reproducibility of results therefore remains an important goal for improving scientific research standards.

Second, incorrect claims made in model info sheets might provide false assurances to reviewers about the correctness of the claims made in a paper. However, by requiring authors to precisely state details about their modeling process, model info sheets enable incorrect claims to be challenged more directly than in status quo, where details about the modeling process are often left undisclosed.

Filling out and evaluating model info sheets requires some expertise in ML. In fields where both authors and reviewers lack any ML expertise, subtle cases of leakage might slip under the radar despite the use of model info sheets. In such cases, we hope that model info sheets released publicly along with papers will enable discourse within scientific communities on the shortcomings of scientific models.

Finally, we acknowledge that our understanding of leakage may evolve, and model info sheets may need to evolve with it. To that end, we have versioned model info sheets, and plan to update them as we continue to better understand leakage in ML-based science.

\section{A case study of civil war prediction} \label{sec:civil-war}

\begin{figure}[b]
    \centering
    \includegraphics[width=\columnwidth]{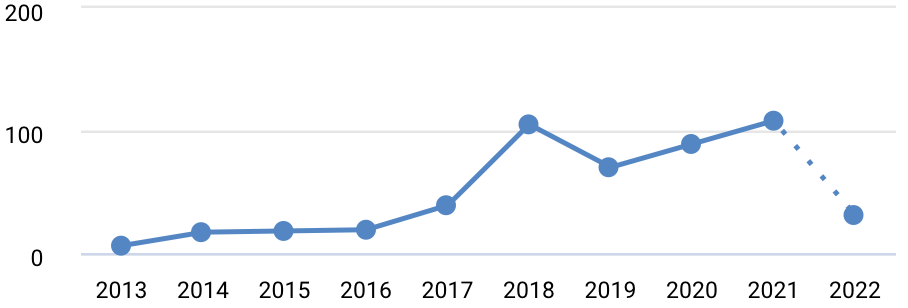}
    \vspace{-0.5cm}
    \caption{Number of political science papers containing the terms \textit{"civil war" AND "machine learning"} in the dimensions database of academic research~\cite{hook_dimensions_2018}. Note the sharp increase in papers using ML methods in the last few years.}
    \label{fig:cwp-trend}
\end{figure}

To understand the impact of data leakage and the efficacy of model info sheets in addressing it, we undertake a reproducibility study in a field where ML models are believed to vastly outperform older statistical models such as Logistic Regression (LR) for predictive modeling: civil war prediction \cite{bara_forecasting_2020}. Over the last few years, this field has switched to predictive modeling using complex ML models such as Random Forests and Adaboost instead of LR (see \cref{fig:cwp-trend}), with several papers claiming near-perfect performance of these models for civil war prediction \cite{muchlinski_comparing_2016, colaresi_robot_2017, wang_comparing_2019, kaufman_improving_2019}.
While the literature we reviewed in our survey highlighted the pitfalls in adopting ML methods (\cref{table:survey}), we go further than most previous research to investigate whether the claims made in the reviewed studies survive once the errors are corrected.

\textbf{Systematic search of predictive modeling literature in civil war research.}
We conducted a systematic search to find relevant literature (detailed in \cref{sec:paper_selection}). This yielded 124 papers. We narrowed this list to the 12 papers that focused on predicting civil war, evaluated performance using a train-test split, and shared the complete code and data. 
For these 12, we attempted to identify errors and reproducibility issues from the text and through reviewing the code provided with the papers. When we identified errors, we re-analyzed the data with the errors corrected.

\textbf{Finding 1: Data leakage causes irreproducible results.} We present our results in \cref{fig:allFive}. We found errors in 4 of the 12 papers---exactly the 4 papers that claimed superior performance of complex ML models over baseline LR models for predicting civil war. All papers suffer from different forms of leakage. All 4 papers were published in top-10 journals in the field of ``Political Science and International Relations'' \cite{scimago_country_2020}. 
When the errors are corrected, complex ML models perform no better than baseline LR models in each case except \citet{wang_comparing_2019}, where the difference between the AUC of the complex ML models and LR models drops from 0.14 to 0.01. 
Further, while none of these errors could have been caught by reading the paper, model info sheets enable the detection of leakage in each case (\cref{app:model_cards}). 
Beyond reproducibility, our results show that complex ML models are not substantively better at civil war prediction than decades old LR models. This is consistent with similar sobering findings in other tasks involving predicting social outcomes such as children’s life outcomes \cite{salganik_measuring_2020} and recidivism \cite{dressel2018accuracy}. Our findings strongly suggest the need for tempering the optimism about predictive modeling in the field of civil war prediction and question the use of ML models in this field.
We provide a detailed overview of our methodology for correcting the errors and show that our results hold under several robustness checks in \cref{sec:corrections}.

\textbf{Finding 2: No significance testing or uncertainty quantification.}
We found that 9 of the 12 papers for which complete code and data were available included no significance tests or uncertainty quantification for classifier performance comparison (\cref{tab:significance}). 
Especially when sample sizes are small, significance testing and uncertainty quantification are important steps towards reproducibility 
\cite{mcdermott_reproducibility_2021, gorman_we_2019}.
As an illustration, we examine this issue in detail in the case of \citet{blair_forecasting_2020} since their test dataset has a particularly small number of instances of civil war onset (only 11). They propose a model of civil war onset that uses theoretically informed features and report that it outperforms other baseline models of civil war onset using the AUC metric on an out-of-sample dataset. 
We find that the performance of their model is not significantly better than other baseline models for civil war prediction.\footnote{\textit{Z} = 0.64, 1.09, 0.42, 0.67; \textit{p} = 0.26, 0.14, 0.34, 0.25 for a one-tailed significance test comparing the smoothed AUC performance of the model proposed in the paper---the \textit{escalation} model---with other baseline models reported in their paper---\textit{quad, goldstein, cameo} and \textit{average} respectively. 
We implement the comparison test for smoothed ROC curves detailed by Robin et al. \cite{robin_proc_2011}. Note that we do not correct for multiple comparisons; such a correction would further reduce the significance of the results.}
Further, all models have large confidence intervals for their out-of-sample performance. For instance, while the smoothed AUC performance reported by the authors is 0.85, the 95\% confidence interval calculated using bootstrapped test set re-sampling is [0.66-0.95].

\section{Beyond leakage: enhancing the reproducibility of ML-based science}
\label{sec:discussion}

We have outlined how the use of model info sheets can address data leakage in ML-based science. In addition to leakage, we found a number of other reproducibility issues in our survey. Here, we present five diagnoses for reproducibility failures in fields adopting ML methods. Each of our diagnoses is paired with a recommendation to address it.

\textbf{[D1] Lack of understanding of the limits to prediction.} Recent research for predicting social outcomes has shown that even with complex models and large datasets, there are strong limits to predictive performance \cite{salganik_measuring_2020,dressel2018accuracy}. However, results like the better-than-human performance of ML models in perception tasks such as image classification \cite{he_delving_2015, szeliski2021computer} give the impression of ML models surpassing human performance across tasks, which can confuse researchers about the performance they should realistically expect from ML models.

\textbf{[R1] Understand and communicate limits to prediction.} A research agenda which investigates the efficacy of ML models in tasks across scientific fields would increase our understanding of the limits to prediction. This can alleviate the overoptimism that arises from confusing progress in one task (e.g., image classification) with another (e.g., predicting social outcomes). If we can identify upper bounds on the predictive accuracy of tasks (i.e., lower bound the Bayes Error Rate for a task), then once the achievable accuracy has been reached, we can avoid a futile effort to increase it further and can apply increased skepticism towards results that claim to violate known bounds.

\textbf{[D2] Hype, overoptimism and publication biases.} 
The hype about commercial AI applications can spill over into ML-based science, leading to overoptimism about their performance. 
Non-replicable findings are cited more than replicable ones \cite{serra-garcia_nonreplicable_2021}, which can result in feedback loops of overoptimism in ML-based science.
Besides, publication biases that have been documented in several scientific fields \cite{shi_trim-and-fill_2019, gurevitch_meta-analysis_2018} can also affect ML-based science \cite{hofman_prediction_2017, islam_reproducibility_2017}.

\textbf{[R2] Treat results from ML-based science as tentative.} When overoptimism is prevalent in a field, it is important to engage with results emerging from the field critically. Until reproducibility issues in ML-based science are widely addressed and resolved, results from this body of work should be treated with caution.

\textbf{[D3] Inadequate expertise.} The rapid adoption of ML methods in a scientific field can lead to errors. These can be caused due to the lack of expertise of domain experts in using ML methods and vice-versa. 

\textbf{[R3] Inter-disciplinary collaborations and communication of best-practices.} Literature in the ML community should address the different failure modes that arise during the modeling process. 
Researchers with expertise in ML methods should clearly communicate best practices in deploying ML for scientific research \cite{lones_how_2021}. Having an interdisciplinary team consisting of researchers with domain expertise as well as ML expertise can avoid errors.

\textbf{[D4] Lack of standardization.}
Several applied ML fields, such as engineering applications and modeling contests, have adopted practices such as standardized train-test splits, evaluation metrics, and modeling tasks to ensure the validity of the modeling and evaluation process \cite{russakovsky_imagenet_2015, koh_wilds_2021}. However, many of these have not yet been adopted widely in ML-based science. This leads to subtle errors in the modeling process that can be hard to detect. 

\textbf{[R4] Adopt the common task framework when possible.} 
The common task framework allows us to compare the performance of competing ML models using an agreed-upon training dataset and evaluation metrics, a secret holdout dataset, and a public leaderboard \cite{rocca_putting_2021, donoho_50_2017}. Dataset creation and model evaluation is left to impartial third parties who have the expertise and incentives to avoid errors.
However, one undesirable outcome that has been observed in communities that have adopted the common task framework is a singular focus on optimizing a particular accuracy metric to the exclusion of other scientific and normatively desirable properties of models \cite{paullada2020data, marie_scientific_2021, gorman_we_2019}.

\textbf{[D5] Lack of computational reproducibility.}
The lack of computational reproducibility hinders verification of results by independent researchers (\cref{subsec:other_issues}). While computational reproducibility does not mean that the code is error-free, it can make the process of finding errors easier, since researchers attempting to reproduce results do not have to spend time getting the code to run. 

\textbf{[R5] Ensure computational reproducibility.} 
Platforms such as CodeOcean \cite{clyburne_computational_2019}, a cloud computing platform which replicates the exact computational environment used to create the original results, can be used to ensure the long term reproducibility of results. 
We follow several academic journals and researchers in recommending that future research in fields using ML methods should use similar methods to ensure computational reproducibility \cite{noauthor_easing_2018, liu_successes_2019}.

\section{Conclusion}

The attractiveness of adopting ML methods in scientific research is in part due to the widespread availability of off-the-shelf tools to create models without expertise in ML methods \cite{hutson_no_2019}. However, this \textit{laissez faire} approach leads to common pitfalls spreading to all scientific fields that use ML. So far, each research community has independently rediscovered these pitfalls. 
Without fundamental changes to research and reporting practices, we risk losing public trust owing to the severity and prevalence of the reproducibility crisis across disciplines. 
Our paper is a call for interdisciplinary efforts to address the  crisis by developing and driving the adoption of best practices for ML-based science. Model info sheets for detecting and preventing leakage are a first step in that direction.

\textbf{Materials and methods.}
The code and data required to reproduce our case study on civil war prediction is uploaded to a CodeOcean capsule (\url{https://doi.org/10.24433/CO.4899453.v1}). \cref{sec:corrections} contains a detailed description of our methods and results from additional robustness checks.

\textbf{Acknowledgements.} We are grateful to Jessica Hullman, Matthew J. Salganik and Brandon Stewart for their valuable feedback on drafts of this paper. We thank Robert Blair, Aaron Kaufman, David Muchlinski and Yu Wang for quick and helpful responses to drafts of this paper. We are especially thankful to Matthew Sun, who reviewed our code and provided helpful suggestions and corrections for ensuring the computational reproducibility of our own results, and to Angelina Wang, Orestis Papakyriakopolous, and Anne Kohlbrenner for their feedback on model info sheets.

\bibliography{references1,references-arvind, references-appendix}
\bibliographystyle{icml2022}

\newpage
\appendix
\onecolumn
\section*{Appendix}

\setcounter{table}{0}
\renewcommand{\thetable}{A\arabic{table}}
\setcounter{figure}{0}
\renewcommand{\thefigure}{A\arabic{figure}}

\paragraph{Overview of the Appendix.} 

In \cref{app:terminology}, we justify our choice of the word reproducibility.
In \cref{sec:corrections}, we provide a detailed description of the methods we used to select papers for our review of civil war prediction and fix reproducibility issues in the papers with errors. 
In \cref{app:model_cards}, we show that each type of leakage identified in our survey (\cref{sec:taxonomy}) is addressed by model info sheets. 
We provide a template of model info sheets and a list of all 124 papers that we considered for our literature review in civil war prediction on our website (\url{https://reproducible.cs.princeton.edu}).

\pgfplotsset{compat=1.9}%

\pgfkeys{/pgf/fpu=true}

\section{Why do we call these reproducibility issues?} \label{app:terminology}

We acknowledge that there isn't consensus about the term reproducibility, and there have been a number of recent attempts to define the term and create consensus \cite{national_academies_of_sciences_reproducibility_2019}. One possible definition is computational reproducibility---when the results in a paper can be replicated using the exact code and dataset provided by the authors \cite{liu_successes_2019}. We argue that this definition is too narrow because even cases of outright bugs in the code would not be considered irreproducible under this definition. Therefore we advocate for a standard where bugs and other errors in data analysis that change or challenge a paper's findings constitute irreproducibility. 

The goal of predictive modeling is to estimate (and improve) the accuracy of predictions that one might make in a real-world scenario. This is true regardless of the specific research question one wishes to study by building a predictive model. In practice one sets up the data analysis to mimic this real-world scenario as closely as possible. There are limits to how well we can do this and consequently there is always methodological debate on some issues, but there are also some clear rules. If an analysis choice can be shown to lead to incorrect estimates of predictive accuracy, there is usually consensus in the ML community that it is an error. For example, violating the train-test split (or the learn-predict separation) is an error because the test set is intended to provide an accurate estimate of 'out-of-sample' performance---model performance on a dataset that was not used for training \cite{kuhn_applied_2013}. Thus, to define what is an error, we look to this consensus in the ML community (e.g. in textbooks) and offer our own arguments when necessary.

\section{Materials and Methods: Reproducibility issues in civil war prediction}
\label{sec:corrections}

Different researchers might have different aims when comparing the performance on civil war prediction --- determining the absolute performance, or comparing the relative performance of different models of civil war prediction. Whether the aim is to determine the relative or absolute performance of models of civil war prediction, data leakage causes a deeper issue in the findings of each of the 4 papers with errors that leads to inaccurate estimates of both relative and absolute out-of-sample performance.

In correcting the papers with errors \cite{muchlinski_comparing_2016, colaresi_robot_2017, wang_comparing_2019, kaufman_improving_2019}, our aim is to report out-of-sample performance of the various models of civil war prediction after correcting the data leakage, while keeping all other factors as close to the original implementation as possible. 
Fixing the errors allows a more accurate estimate of out-of-sample performance.

At the same time, we caution that just because our corrected results offer a more accurate estimate of out-of-sample performance doesn't mean that we endorse all other methodological choices made in the papers. 
For example, to correct the results reported by \citet{muchlinski_comparing_2016}, we use imputation on an out-of-sample dataset that has 95\% missing values. While an imputation model created only using the training data avoids data leakage, it does not mean that using a dataset with 95\% missing values to measure out-of-sample performance is desirable.

\subsection{Paper selection for review}\label{sec:paper_selection}
To find relevant papers on civil war prediction for our review, we used the search results from a dataset of academic literature \cite{hook_dimensions_2018} for papers with the terms \textit{`civil' AND `war' AND (`prediction' OR `predicting' OR `forecast')} in their title or abstract, as well as papers that were cited in a recent review of the field \cite{bara_forecasting_2020}. 
To keep the number of papers tractable, we limited ourselves to those that were published in the last 5 years, specifically, papers published between 1st January 2016 and 14th May 2021. 
This yielded  124 papers. We narrowed this list to the 15 papers that were focused on predicting civil war and evaluated performance using a train-test split.
Of the 15 papers that meet our inclusion criteria, 12 share the complete code and data. For these 12, we attempted to identify errors and reproducibility issues from the text and through reviewing the code provided with the papers. When we identified errors, we re-analyzed the data with the errors corrected.
We now address the reproducibility issues we found in each paper in detail.

\subsection{\citet{muchlinski_comparing_2016}}

Imputation is commonly used to fill in missing values in datasets \cite{donders_review_2006}. 
Imputing the training and test datasets together refers to using data from the training as well as the test datasets to create an imputation model that fills in all missing values in the dataset. 
This is an erroneous imputation method for the predictive modeling paradigm, since it can lead to data leakage, which results in incorrect, over-optimistic performance claims. 
This pitfall is well known in the predictive modeling community --- discussed in ML textbooks \cite{kuhn_applied_2013}, blogs \cite{ghani_top_2020} and popular online forums~\cite{noauthor_imputation_nodate}.

\citet{muchlinski_comparing_2016} claim that a Random Forests model vastly outperforms Logistic Regression models in terms of out-of-sample performance using the AUC metric \cite{fawcett_introduction_2006}. However, since they impute the training and test datasets together, their results suffer from data leakage. 
The impact of leakage is especially severe because of the level of missingness in their out-of-sample test dataset: over 95\% of the values are missing (which is not reported in the paper), and 70 of the 90 variables used in their model are missing for {\em all} instances in the out-of-sample test set.\footnote{While leakage is particularly serious in predictive modeling, a dataset with 95\% of values missing is problematic even for explanatory modeling.} 
When their imputation method is corrected, their Random Forests model performs no better than the  Logistic Regression models that they compared against.

We focus on reproducing the out-of-sample results reported by \citet{muchlinski_comparing_2016}. \Cref{tab:muchlinski-comparison} provides the comparisons between the results reported in \citet{muchlinski_comparing_2016}, our reproductions of their reported (incorrect) results, as well as the corrected version of their results. 
\citet{muchlinski_comparing_2016} received two critiques of the methods used in their paper \cite{wang_comparing_2019, neunhoeffer_how_2019}. \footnote{\citet{hofman_expanding_2021} also outline the shortcomings in the initial code released by \citet{muchlinski_comparing_2016}.}.
In response, they published a reply with clarifications and revised code addressing both critiques \cite{muchlinski_seeing_2019}. 
We use the revised version of their code. 
We find that the error in their imputation methods exists in the revised code as well as the original code, and was not identified by the previous critiques. 
\citet{muchlinski_comparing_2016} re-use the dataset from \citet{hegre_sensitivity_2006} when training their models, and provide a separate out-of-sample test set for evaluation. To address missing values, they use a Random Forests based imputation method in R called \textit{rfImpute}.
However, the training and test sets are imputed together, which leads to a data leakage. This results in overoptimistic performance claims. Below, we detail the steps we take to correct their results, provide a visualization of the data leakage, and provide a simulation showcasing how the data leakage can result in overoptimistic claims of performance.

\paragraph{Correcting the data imputation.} To correct this error, we use the \textit{mice} package in R which uses multiple imputation for imputing missing data. This is because the \textit{mice} package allows us to specify which rows in the dataset are a part of the test set and it does not use those rows for creating the imputation model, whereas \textit{rfImpute} --- the original method used to impute the missing data in the original results by \citet{muchlinski_comparing_2016} --- does not have this feature. The authors imputed the training set together with the out-of-sample test set using \textit{rfImpute}, which led to data leakage. \Cref{tab:muchlinski-comparison} provides the comparisons between the results reported in \citet{muchlinski_comparing_2016}, our reproductions of their reported (incorrect) results, as well as the corrected version of their results.

Using multiple imputation fills in missing values without regarding the underlying variable's original distribution. For example, using multiple imputation fills in different missing values for the variable representing the percentage of rough terrain in a country in different years \cite{beger_tweet_2021}, whereas this particular variable (percentage of rough terrain) is constant over time. However, when multiple imputation is used with a train-test split, there is still no leakage between the training and test sets, since the imputation model only uses data from the training set to fill in missing values in the test set.

\paragraph{Why can't we use \textit{rfImpute} in the corrected results?} Instead of using the \textit{mice} package, another way to impute the data correctly, i.e., without data leakage, would be to run the imputation using \textit{rfImpute} on the training and test data separately --- creating two separate imputation models --- one for the training data and one for the test data. We could not use this imputation method because 70 of the 90 variables used in \citet{muchlinski_comparing_2016}'s model as features do not have \textit{any} values in the out-of-sample test data provided --- i.e. they are missing for \textit{all} observations in the out-of-sample dataset --- and \textit{rfImpute} requires at least some values for each variable to not be missing. In other words, the \textit{mice} package allows us to train an imputation model on the training set and use it to fill in missing values in the test set. 

\paragraph{Subtle differences between explanatory and predictive modeling.} In the explanatory modeling paradigm, the aim is to draw inferences from data, as opposed to optimizing and evaluating out-of-sample predictive performance. In this case, data imputation would be considered a part of the data pre-processing step, even though it is still important to keep in mind the various assumptions being made in this process \citet{schafer_multiple_1999}. Contrarily, in the predictive modeling paradigm, the imputation is a part of the modeling step \cite{kuhn_applied_2013} because the aim of the modeling exercise is to validate performance on an out-of-sample test set, which the model does not have access to during the training. In this case, imputing the training and test datasets together leads to leaking information from the test set to the training set and thus the performance evaluation on the purportedly ``out-of-sample'' test set would be an over-estimate. 

\paragraph{What is the precise mechanism by which the leakage occurs in \citet{muchlinski_comparing_2016}?} When \citet{muchlinski_comparing_2016} impute the missing values in the out-of-sample test set, the imputation model has access to the entire training data as well as the labels of the target variables in the test data --- they also include the target variable in the list of variables which the imputation model treats as independent variables when carrying out the imputation. The model therefore uses correlations between the target variable and independent variables in the training dataset and uses them to fill in the missing values in the test dataset --- i.e. the model uses the labels of the target variables in the test data and correlations from the training data to fill in missing values. This leads to the test dataset having similar correlations between the target and independent variables as the ones present in the training data. 
Further, the missing data is filled in in such a way that it favors ML models such as Random Forests over Logistic Regression models, as we show in the visualization below.

\begin{figure}
    \centering
    \begin{subfigure}[b]{0.4\textwidth}
        \centering
        \includegraphics[width=\textwidth]{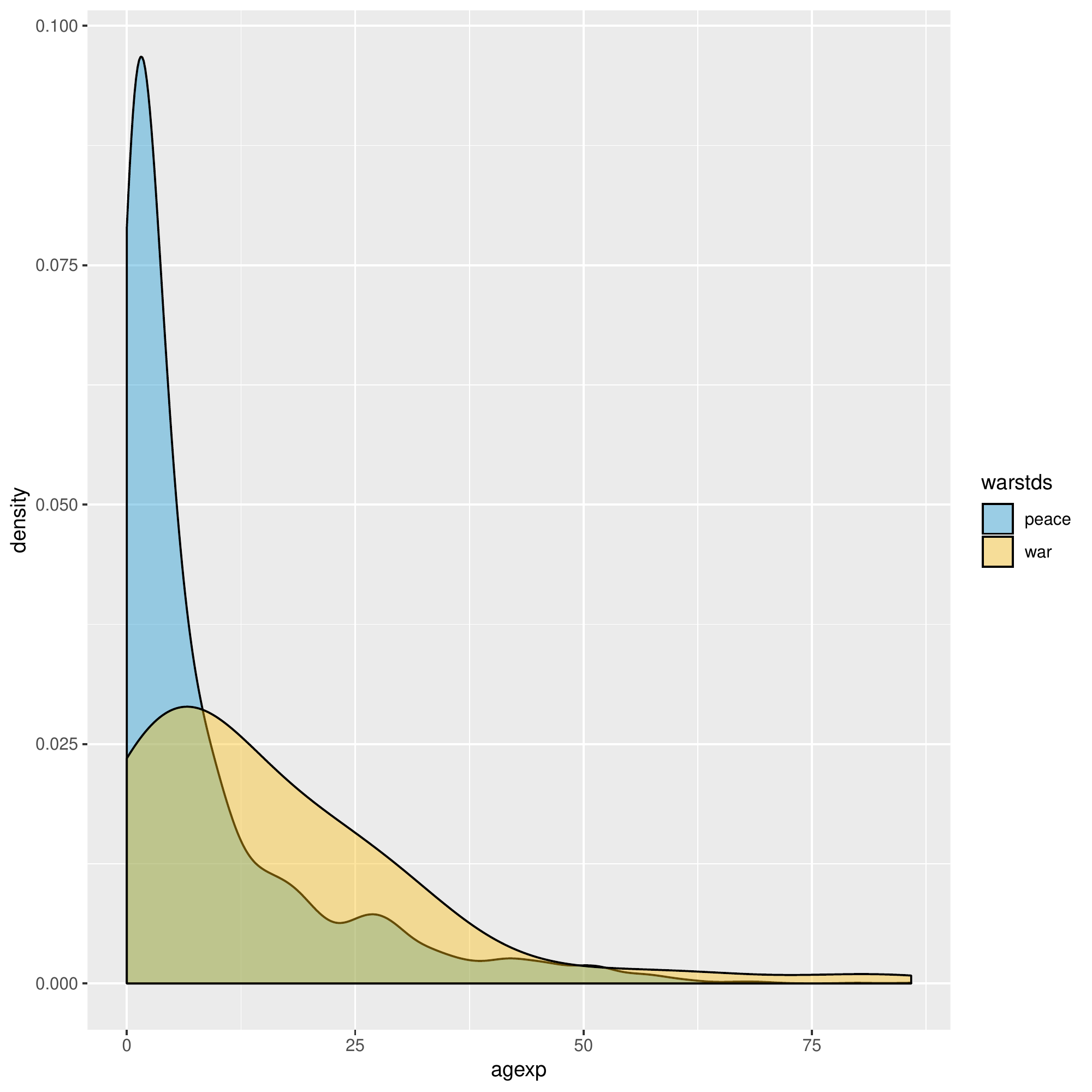}
        \caption{Distribution of the \textit{agexp} variable for peace and war data points for the original Hegre et al. dataset, ignoring missing values}  
        \label{fig:much-orig}
    \end{subfigure}
    \hfill
    \begin{subfigure}[b]{0.4\textwidth}  
        \centering 
        \includegraphics[width=\textwidth]{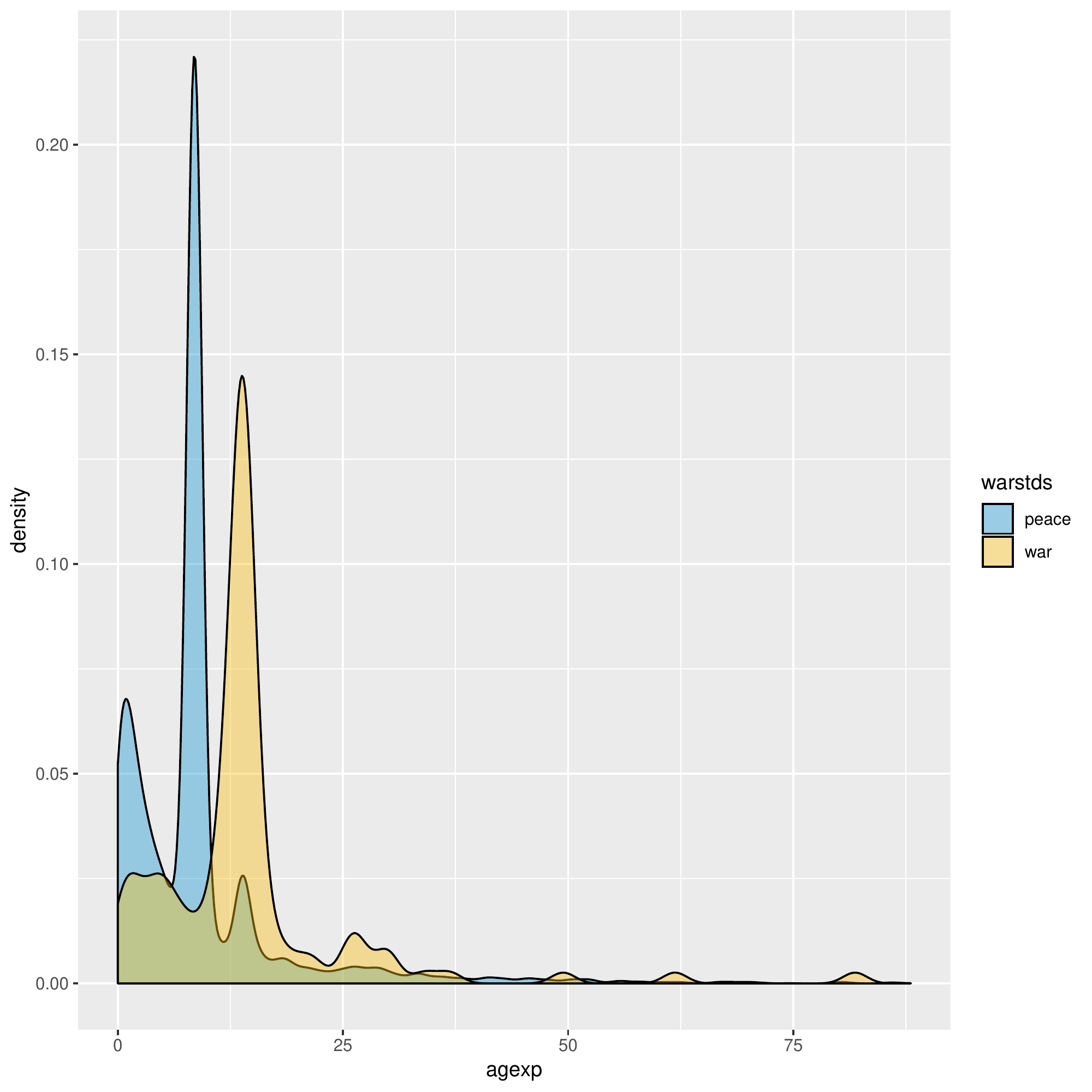}
        \caption{Distribution of the \textit{agexp} variable for peace and war data points for the imputed Hegre et al. dataset used by Muchlinski et al. for training}   
        \label{fig:much-orig-imp}
    \end{subfigure}
    \hfill
    \begin{subfigure}[b]{0.4\textwidth}   
        \centering 
        \includegraphics[width=\textwidth]{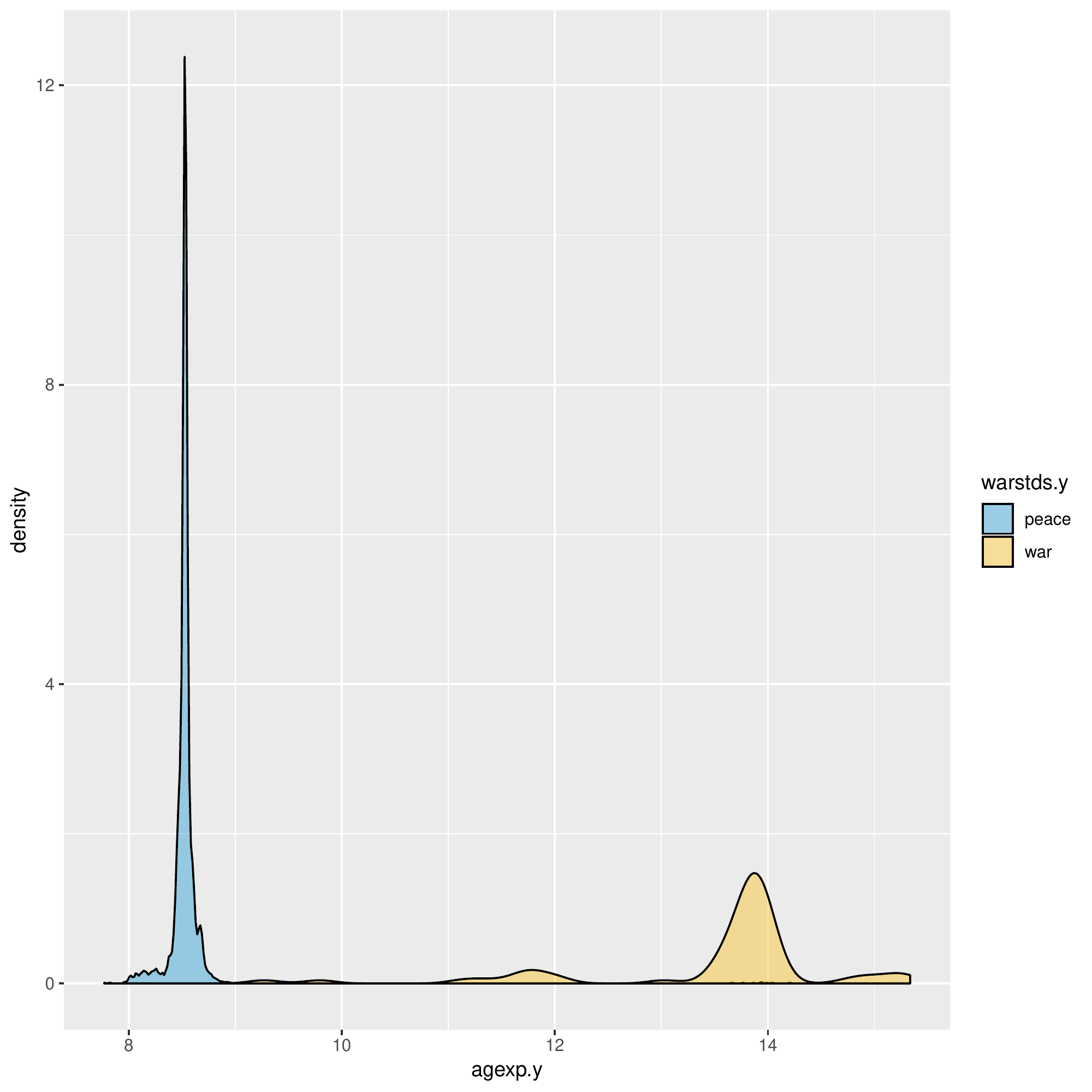}
        \caption{Distribution of the \textit{agexp} variable for peace and war data points only for the data points that were added during imputation (i.e. the data points that were missing in the original dataset)}   
        \label{fig:much-imp}
    \end{subfigure}
    \hfill
    \begin{subfigure}[b]{0.4\textwidth}   
        \centering 
        \includegraphics[width=\textwidth]{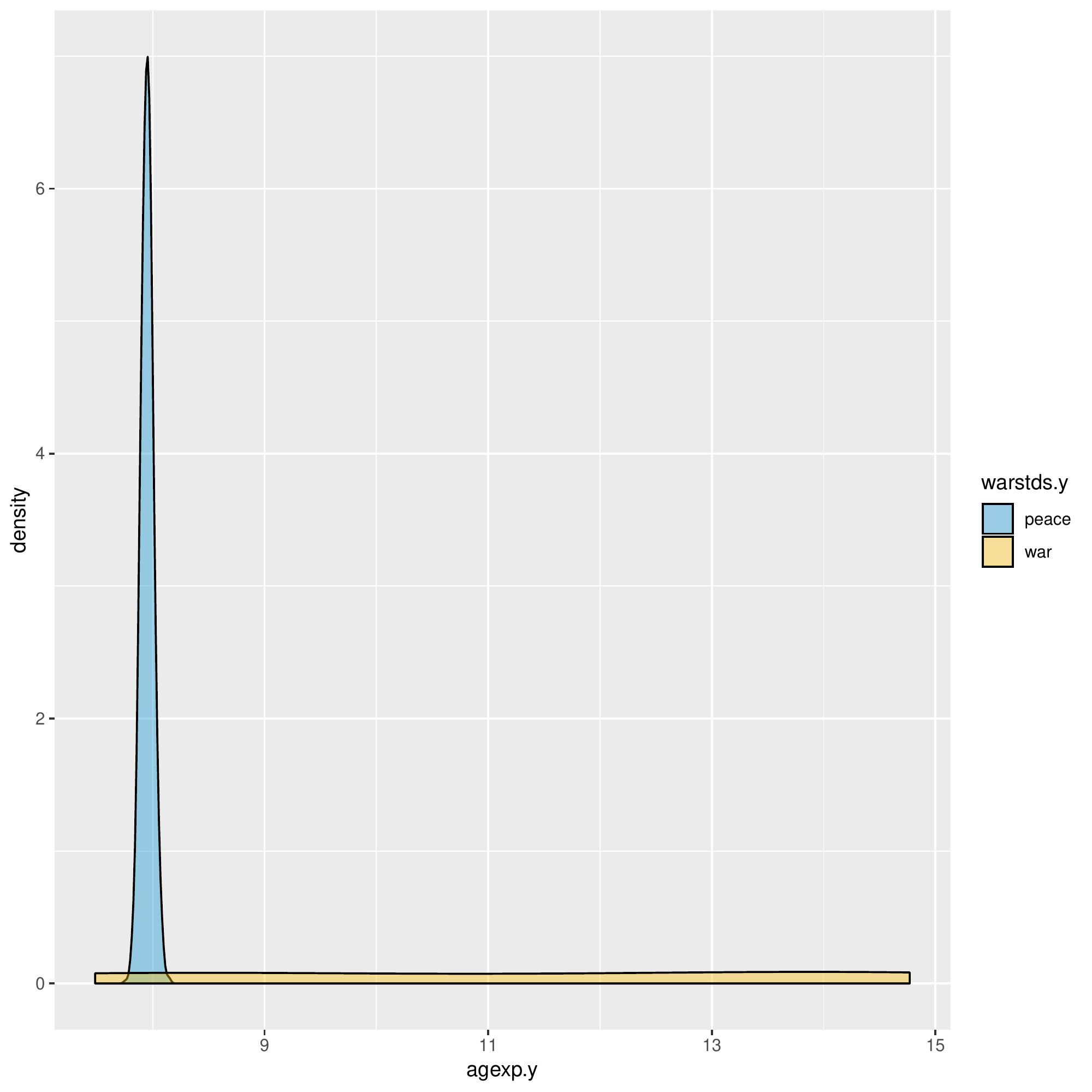}
        \caption{Distribution of the \textit{agexp} variable for peace and war data points for the out-of-sample test set}   
        \label{fig:much-oos}
    \end{subfigure}
    \caption{Distribution of the \textit{agexp} variable for peace and war data points for different imputation steps in \citet{muchlinski_comparing_2016}. Note that the distribution of \textit{peace} instances in the test set (D) has a peak that is close to the distribution in the imputed training set (B, C) --- which allows the random forests model to learn the small range of values where \textit{peace} data points are concentrated. While we report results for the \textit{agexp} variable, similar trends appear across independent variables in the dataset.}
    \label{fig:much-agexp}
\end{figure}

\paragraph{Visualizing the leakage.} We can visually observe an instance of data leakage in \Cref{fig:much-agexp}. We focus on the distribution of the feature \textit{agexp}, which represents the proportion of agricultural exports in the GDP of a country. We choose this feature because in the Muchlinski et al. paper, this feature had the highest gini index for the random forests model --- which means that it was an important feature for the model. While we only visualize one feature here, similar results hold across multiple features used in the model. Below, we reconstruct the process by which the data leakage was generated --- following the exact steps \citet{muchlinski_comparing_2016} used to create and evaluate the dataset:

\begin{itemize}
    \item \Cref{fig:much-orig} represents the distribution of the \textit{agexp} variable for war and peace data points in the original dataset by \citet{hegre_sensitivity_2006}, ignoring missing values. 
    \item \Cref{fig:much-orig-imp} shows the same distribution after including the imputed values of \textit{agexp}. In particular, we see two peaks in the dataset for war and peace data points alike, one due to war instances and one due to peace instances. 
    \item If we look only at the data points that were imputed using the \textit{rfImpute} method (\Cref{fig:much-imp}), we see that the distribution of the imputed data points for war and peace are completely separated, in contrast to the original distribution where there was a significant overlap between the distributions. 
    \item Finally, \Cref{fig:much-oos} shows the effect of imputing this already-imputed dataset with the out-of-sample test set --- we see that the out-of-sample dataset only has the peak for peace datapoints, whereas the distribution for war is almost uniform.  
\end{itemize}

Further, the random forests model can learn the peak for the \textit{agexp} variable in the \textit{peace} instances from the training dataset after imputation, since the peak for the training and test sets is similar. It can distinguish between war and peace datapoints much more easily compared to a logistic regression model that only uses one parameter per feature — logistic regression models are monotonic functions of the independent variables and therefore cannot learn that a variable only lies within a small range for a given label. This highlights the reason behind Random Forests outperforming Logistic Regression in this setting --- imputing the training and test datasets together leads to variable values being artifically concentrated within a very small range for both the training and test datasets --- and further, being neatly separated across \textit{war} and \textit{peace} instances. The impact of the imputation becomes even clearer when we consider that the out-of-sample test dataset provided by \citet{muchlinski_comparing_2016} has over 95\% of the data missing, and 70 out of 90 variables are missing for all instances in the out-of-sample dataset. 

\begin{figure}[h!]
    \centering
    \includegraphics[width=0.5\textwidth]{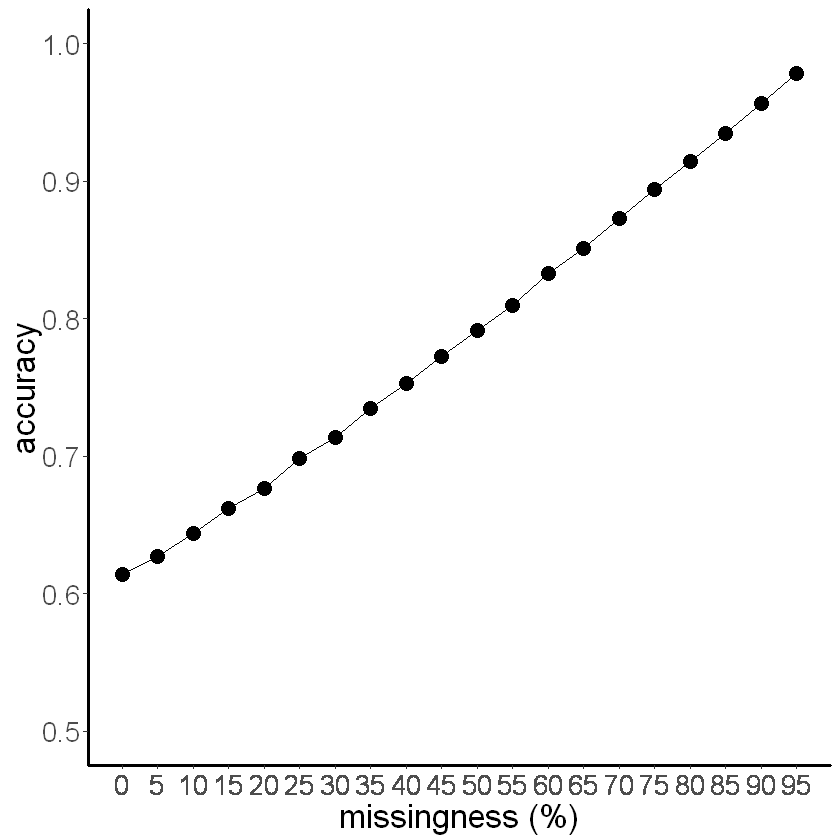}
    \caption{Results of a simulation that showcase how imputing the training and test sets together leads to overoptimistic estimates of model performance. The 95\% Confidence Intervals are too small to be seen. }
    \label{fig:much-simulation}
\end{figure}

\paragraph{A simulation showcasing the impact of missingness on performance estimates in the presence of leakage.} We can observe a visual example of how data leakage affects performance evaluation in \Cref{fig:much-simulation}. We describe the simulation below:

\begin{itemize}
    \item there are two variables --- the target variable \textit{onset} and the independent variable \textit{gdp}. 
    \item \textit{onset} is a binary variable. \textit{gdp} is drawn from a normal distribution and depends on \textit{onset} as follows: $$gdp = N(0,1) + onset.$$
    \item We generate 1000 samples with \textit{onset=0} and 1000 samples with \textit{onset=1} to create the dataset.
    \item We randomly split the data into training (50\%) and test (50\%) sets, and create a random forests model that is trained on the training set and evaluated on the test set. 
    \item To observe the impact of imputing the training and test sets together, we randomly delete a certain percentage of values of \textit{gdp}, and impute it using the imputation method used in \citet{muchlinski_comparing_2016}.
    \item We vary the proportion of missing values from 0\% to 95\% in increments of 5\% and plot the accuracy of the random forests classifier on the test set.
    \item We run the entire process 100 times and report the mean and 95\% CI of the accuracy in \Cref{fig:much-simulation}; the 95\% CI is too small to be seen in the Figure.
\end{itemize}  

We find that imputing the training and test sets together leads to an increasing improvement in the purportedly ``out-of-sample'' accuracy of the model. Estimates of model performance in this case are artificially high. This example also highlights the impact of the high percentage of missing values --- since the out-of-sample test set used by \citet{muchlinski_comparing_2016} contains over 95\% missing values, the impact of imputing the training and test sets together is very high.

\begin{table}
\centering
\pgfplotstabletypeset[
col sep=comma,
columns={Algorithm, Reported, Reported results (reproduced), Corrected results},
columns/Algorithm/.style={column type={c|},string type},
]{muchlinski-comparison.csv}
\caption{Original and corrected results in \citet{muchlinski_comparing_2016}. While there are differences between the reported results and our reproduction of the reported results, especially for the Fearon and Laitin as well as the Collier and Hoeffler models, the relative order of the model performance for both results is the same.}
\label{tab:muchlinski-comparison}
\end{table}

\subsection{\citet{colaresi_robot_2017}}

\citet{colaresi_robot_2017}
report that ML models vastly outperform Logistic Regression for predicting civil war onset.
However, they re-use the imputed version of the dataset in \citet{hegre_sensitivity_2006} which is provided by \citet{muchlinski_comparing_2016}. 
They use the imputed dataset both for training and testing via a train-test split; they do not use the out-of-sample test set provided by Muchlinski et al. 
This means that the results in \citet{colaresi_robot_2017} are subject to exactly the same pitfall as in \citet{muchlinski_comparing_2016}, albeit with a slightly different dataset.
Correcting the imputation method dramatically reduces the performance of the ML models proposed.

We focus on reproducing the final round of results reported in the paper \citet{colaresi_robot_2017}, which consists of a comparison of 3 models of civil war onset --- the Random Forests model proposed in \citet{muchlinski_comparing_2016}, the Random Forests model proposed in \citet{colaresi_robot_2017} as well as the Logistic Regression model proposed in \citet{fearon_ethnicity_2003}. 
Their dataset has 17.4\% values missing, and the test set has 19\% values missing. The proportion of missing values in individual variables can be even higher --- for example, the \textit{agexp}, which represents the proportion of agricultural exports in the GDP of a country, is missing for 54.3\% of the rows in the test set.
In our corrected results, we use the original dataset from \citet{hegre_sensitivity_2006} and impute the training and test data separately using the \textit{rfImpute} function. The test set consists of data from the years after 1988. One of the independent variables, \textit{milper}, is missing for all instances in the test set of \citet{colaresi_robot_2017} so we exclude this variable from our models. \Cref{tab:colaresi-comparison} provides the comparisons between the results reported in \citet{colaresi_robot_2017}, our reproductions of their reported (incorrect) results, as well as the corrected version of their results. 

\citet{colaresi_robot_2017} and \citet{wang_comparing_2019} reuse the dataset released by \citet{muchlinski_comparing_2016}. This is the imputed version of the dataset released by \citet{hegre_sensitivity_2006}.
However, for 777 rows in the imputed dataset released by \citet{muchlinski_comparing_2016}, the original dataset by \citet{hegre_sensitivity_2006} has a missing target variable (i.e. the variable representing civil war onset is missing) whereas the imputed version of the dataset (i.e. the dataset released by \citet{muchlinski_comparing_2016}) has a value of \textit{peace} for the target variable representing civil war onset.
Since \citet{muchlinski_comparing_2016} do not share the code that they use for imputing the \citet{hegre_sensitivity_2006} dataset, it is unclear how the missing values in the target variable were imputed in the dataset, especially since the imputation method they use --- \textit{rfImpute} --- requires non-missing values in the target variable. 
Still, the number of instances of civil war onset (i.e. instances where the variable representing civil war onset has the value \textit{war}) in the \citet{hegre_sensitivity_2006} dataset as well as the \citet{muchlinski_comparing_2016} dataset are the same.

\begin{table}
\centering
\pgfplotstabletypeset[col sep=comma,columns={Algorithm, Reported, Reported results (reproduced), Corrected results}, columns/Algorithm/.style={column type={c|},string type},]{colaresi-comparison.csv}
\caption{Original results from  \citet{colaresi_robot_2017} and our corrected results. }
\label{tab:colaresi-comparison}
\end{table}

\subsection{\citet{wang_comparing_2019}}

Similar to \citet{colaresi_robot_2017}, \citet{wang_comparing_2019} 
report that ML models vastly outperform Logistic Regression for predicting civil war onset.
However, they too re-use the imputed version of the dataset in \citet{hegre_sensitivity_2006} provided by Muchlinski et al. \cite{muchlinski_comparing_2016}. 
They use the imputed dataset both for training and testing via k-fold cross-validation; they do not use the out-of-sample test set provided by Muchlinski et al. 
Correcting the imputation method dramatically reduces the performance of the ML models proposed.

We focus on reproducing the results of the nested cross-validation implementation reported by \citet{wang_comparing_2019}. \citet{wang_comparing_2019} reuses the imputed dataset provided by \citet{muchlinski_comparing_2016}, instead of using the original dataset provided by \citet{hegre_sensitivity_2006} and imputing the training and test sets separately. The dataset has 17.4\% values missing. The proportion of missing values in individual variables can be even higher --- for example, the \textit{agexp}, which represents the proportion of agricultural exports in the GDP of a country, is missing for 49.8\% of the rows in the data set. In our corrected results, we use the original dataset from \citet{hegre_sensitivity_2006} and impute the training and test data separately using the \textit{rfImpute} function within each cross validation fold. This ensures that there is no data leakage between the training and test sets in each fold. \Cref{tab:wang-comparison} provides the comparisons between the results reported in \citet{wang_comparing_2019}, our reproductions of their reported (incorrect) results, as well as the corrected version of their results.

We also conduct an additional robustness analysis in which we use a separate out-of-sample test set instead of $k-$fold cross validation, since using $k-$fold cross validation with temporal data can also lead to leakage across the train-test split. To maintain comparability between the original and corrected results by testing on the same instances of civil war, we continue to use $k-$fold cross validation in the corrected results in \cref{fig:allFive}. We report the results after making this change in \Cref{tab:wang-comparison}. We use the same train-test split as \citet{colaresi_robot_2017} --- \textit{year $<$ 1988} as training data and the rest as test data --- for the out-of-sample test set. The test set consists of data from the years after 1988. One of the independent variables, \textit{milper}, is missing for all instances in the test set of \citet{colaresi_robot_2017} so we exclude this variable from our models.

Note that the imputation method that should be used depends on the exact model deployment scenario, and should mimic it as closely as possible for accurate performance estimates. For example, in some model deployment settings samples for prediction come in one at a time and in some cases they come in batches. In the former setting, imputing the entire test set together may result in overoptimistic performance evaluations as well, since the deployed model doesn't have access to a batch of samples. Our results may thus offer an upper bound on the performance of civil war prediction models in the case of \citet{colaresi_robot_2017} and \citet{wang_comparing_2019}.

\begin{table}
\centering
\begin{adjustbox}{width=\columnwidth,center}
\pgfplotstabletypeset[col sep=comma,columns={Algorithm, Reported, Reported (reproduced), k-fold CV (corrected), Out-of-sample (corrected)}, columns/Algorithm/.style={column type={c|},string type}, columns/Reported/.style={column type={c},string type}]{wang-comparison.csv}
\end{adjustbox}
\caption{Original and corrected results in the \citet{wang_comparing_2019}. We find that using an out-of-sample test set further favors Logistic Regression models over ML models.  The metric for all results is AUC. *These results were not reported using nested cross-validation in \citet{wang_comparing_2019}. In our reproduction of these reported results, we use nested cross-validation, which ensures that we do not get over-estimates of performance.}
\label{tab:wang-comparison}
\end{table}

\subsection{\citet{kaufman_improving_2019}}
We focus on reproducing the results on civil war prediction in \citet{kaufman_improving_2019}. There are several issues in the paper's results. We outline each issue below and provide a comparison of various scenarios in \Cref{tab:kaufman-comparison} that highlight the precise cause of the performance difference between the original and corrected results, and visualize the robustness of our corrected results. We find that even though there are several issues in \citet{kaufman_improving_2019}, the main difference in performance between the original results they report and our corrected results is due to data leakage.

\paragraph{Data leakage due to proxy variables.} The dataset used by \citet{kaufman_improving_2019} has several variables that, if used as independent variables in models of civil war prediction, could cause data leakage, since they are proxies of the outcome variable. \Cref{tab:kaufman-leakage} lists the  variables in the \citet{fearon_ethnicity_2003} dataset that cause leakage. The first 4 rows outline variables that could be affected by civil wars, as outlined in \citet{fearon_ethnicity_2003}. Therefore, following \citet{fearon_ethnicity_2003}, we use lagged versions of these variables in our correction. The other variables in \Cref{tab:kaufman-leakage} are either direct proxies of outcomes of interest or are missing for all instances for civil war.

\paragraph{Parameter selection for the Lasso model.} \citet{kaufman_improving_2019} use an incorrect parameter selection technique when creating their Lasso model that leads to the model always predicting \textit{peace} (i.e. all coefficients of the variables in the model are always zero). We correct this using a standard technique for parameter selection. Instead of choosing model parameters such that the model always predicts \textit{peace}, we use the \textit{cv.glmnet} function in R to choose a suitable value for model parameters based on the training data.

\paragraph{Using $k-$fold cross validation with temporal data.} $k-$fold cross validation shuffles the dataset before it is divided into training and test datasets. 
When the dataset contains temporal data, the training dataset could contain data from a later date than the test dataset because of being shuffled. To maintain comparability between the original and corrected results by testing on the same instances of civil war, we continue to use $k-$fold cross validation in the corrected results in \cref{fig:allFive}. To evaluate out-of-sample performance without using cross-validation, we use a separate train-test split instead of $k-$fold cross validation and report the difference in results for this scenario in the row \textit{Corrected (out-of-sample)} in  \Cref{tab:kaufman-comparison}. We find that there is no substantial difference between the results when using the out-of-sample test set and $k-$fold cross validation --- in each case, none of the models outperforms a baseline that predicts the outcome of the previous year. We use the same train-test split as \citet{colaresi_robot_2017} --- \textit{year $<$ 1988} as training data and the rest as test data.

\paragraph{Replacing missing values with zeros.} \citet{kaufman_improving_2019} replace missing values in their dataset with zeros, instead of imputing the missing data or removing the rows with missing values. This is a methodologically unsound way of dealing with missing data: for example, the models would not be able to discern whether a variable has a value of zero because of missing data or because it was the true value of the variable for that instance. This risks getting underestimates of performance, as opposed to overoptimistic performance claims. As a robustness check, we impute the training and test data separately in each cross-validation fold using the \textit{rfImpute} function in R and report the results in the \textit{Corrected (imputation)} row of \Cref{tab:kaufman-comparison}. We find that the choice of imputation method does not cause a difference in performance, perhaps because only 0.6\% of the values of variables are missing in the dataset.

\paragraph{Choice of cut-offs for calculating accuracy.} Instead of calculating model cutoffs based on the best cutoff in the training set, Kaufman et al. use the distribution of model scores to decide the cutoffs for calculating accuracy. We include robustness results when we change the cutoff selection procedure to choosing the best cutoffs for the training set in the \textit{Corrected (cutoff choice)} row of \Cref{tab:kaufman-leakage}. We find that the choice of cutoff does not impact the main claim --- the performance of the best model is still worse than a baseline that predicts the outcome of the previous year.

\paragraph{Weak Baseline.} \citet{kaufman_improving_2019} compare their results against a baseline model that always predicts \textit{peace}. We find that a baseline that predicts \textit{war} if the outcome of the target variable was civil war in the previous year and predicts \textit{peace} otherwise is a stronger baseline (Accuracy: 97.5\% vs. 86.1\%;
$\chi^2$=633.7, $p=7.836e$-140 using McNemar's test as detailed in \citet{dietterich1998approximate}), and report results against this stronger baseline in \Cref{tab:kaufman-comparison}.

\paragraph{Confusion about the target variable.} \citet{kaufman_improving_2019} use ongoing civil war instead of civil war onset as the target variable in their models. 
While their abstract mentions that the prediction task they attempt is civil war onset prediction, they switch to using the term \textit{civil war incidence} in later sections, without formally defining this term. To attempt to determine what they mean by this term, we looked at the papers they cite; one of them has the term \textit{civil war incidence} in the title \citet{collier_incidence_2002}, and defines civil war incidence as `observations [that] experienced a start of a civil war'. At the same time, in the introduction, they state that they are `predicting whether civil war occurs in a country in a given year' --- which refers to ongoing civil war instead of civil war onset. This might confuse a reader about the specific prediction task they undertake.

\begin{table}[h!]
\centering
\begin{adjustbox}{width=\columnwidth,center}
\pgfplotstabletypeset[
col sep=comma,
columns={Scenario, ADT, RF, SVM, ERF, Lasso, LR, Baseline, Stronger Baseline}, 
    columns/Scenario/.style={column type={c|},string type}, precision=3,
]{kaufman-comparison.csv}
\end{adjustbox}
\caption{Results for the various scenarios in \citet{kaufman_improving_2019}. We report results up to 3 significant figures in this table because the small difference in performance between AdaBoost and Logistic Regression that is ascribed signifance in \citet{kaufman_improving_2019} can only be observed in the third decimal point. The first 2 values of `Stronger Baseline' are reported as 0 because this baseline was not included in the results of \citet{kaufman_improving_2019}.}
\label{tab:kaufman-comparison}
\end{table}

\begin{table}[h!]
\centering
\begin{adjustbox}{width=\columnwidth,center}
\pgfplotstabletypeset[
    col sep=comma,
    columns={Variable name,Reason for leakage,Variable definition in data documentation}, 
    columns/Variable name/.style={column type={c|},string type},
    columns/Reason for leakage/.style={column type={c},string type},
    columns/Variable definition in data documentation/.style={column type={c},string type},
    ]{kaufman-leakage.csv}
\end{adjustbox}
\caption{This table highlights the variables included as independent variables in \citet{kaufman_improving_2019} which cause a data leakage. In the original use of the dataset, \citet{fearon_ethnicity_2003} include lagged versions of the first 4 variables in the list as independent variables in their model to avoid leakage. Following their use of lagged versions of these variables, we do the same in our correction to avoid leakage. The other variables are proxies for the outcomes of interest and hence we remove them from the models to avoid data leakage.}
\label{tab:kaufman-leakage}
\end{table}

\subsection{\citet{blair_forecasting_2020}}

\begin{figure}[h!]
     \centering
     \begin{subfigure}{0.45\textwidth}
         \centering
         \includegraphics[width=\textwidth]{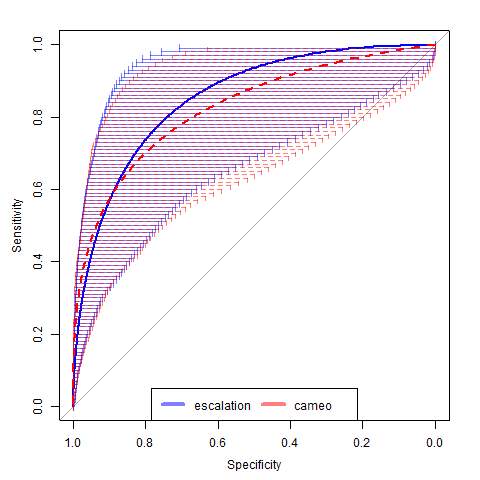}
         \caption{Visualizing the 95\% confidence intervals of the specificities for the 1 month forecast in the smoothed ROC curve reported in \citet{blair_forecasting_2020}.}
     \end{subfigure}
     \hfill
     \begin{subfigure}{0.45\textwidth}
         \centering
         \includegraphics[width=\textwidth]{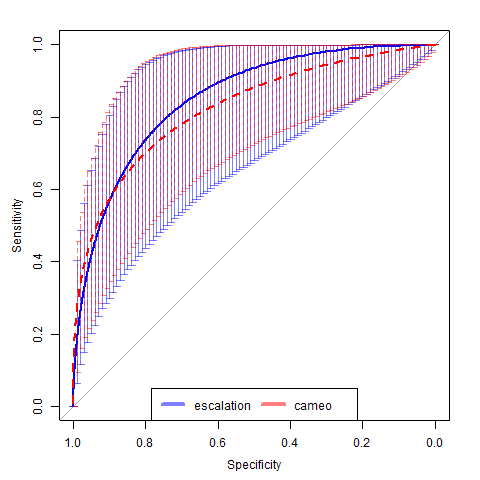}
         \caption{Visualizing the 95\% confidence intervals of the sensitivities for the 1 month forecast in the smoothed ROC curve reported in \citet{blair_forecasting_2020}.}
     \end{subfigure}
        \vskip\baselineskip
    \begin{subfigure}{0.45\textwidth}
         \centering
         \includegraphics[width=\textwidth]{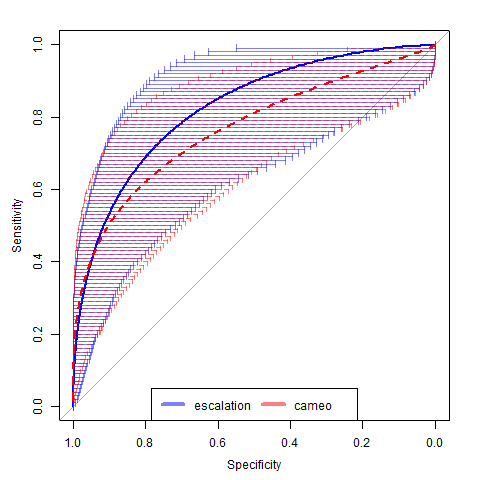}
         \caption{Visualizing the 95\% confidence intervals of the specificities for the 6 month forecast in the smoothed ROC curve reported in \citet{blair_forecasting_2020}.}
     \end{subfigure}
     \hfill
     \begin{subfigure}{0.45\textwidth}
         \centering
         \includegraphics[width=\textwidth]{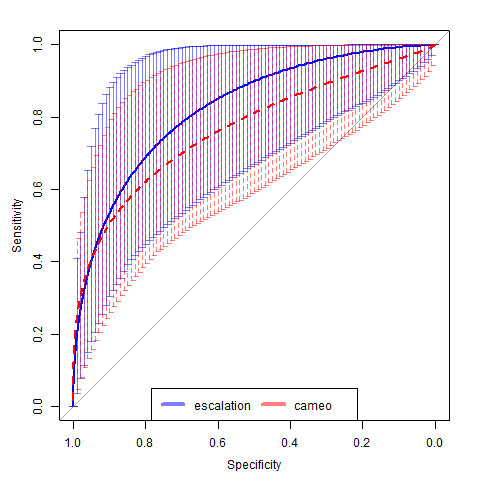}
         \caption{Visualizing the 95\% confidence intervals of the sensitivities for the 6 month forecast in the smoothed ROC curve reported in \citet{blair_forecasting_2020}.}
     \end{subfigure}
        \caption{The wide confidence intervals for sensitivities and specificities reported in Blair and Sambanis. Here, we visualize the \textit{escalation} and \textit{cameo} models for the 1 month and 6 month forecast in the base specification (reported in Figure 1 of their paper). }
        \label{fig:blair-conf}
\end{figure}

\citet{blair_forecasting_2020} state that their \textit{escalation} model outperforms other models across a variety of settings. However, they do not test the performance evaluations to see if the difference is statistically significant. We find that there is no significant difference between the smoothed AUC values of the \textit{escalation} model's performance and other models they compare it to when we use a test for significance. Further, we provide a visualization of the 95\% confidence intervals of specificities and sensitivities in the smoothed ROC curve they report for their model (\textit{escalation}) as well as for a baseline model (\textit{cameo}) --- and find that the 95\% confidence intervals are large (see \Cref{fig:blair-conf}).

\paragraph{Uncertainty quantification, p-values and  Z-values for tests of statistical significance.} 

\begin{itemize}
    \item We report p-values and Z values for a one-tailed significance test comparing the smoothed AUC performance of the \textit{escalation} model with other baseline models reported in their paper --- \textit{quad, goldstein, cameo} and \textit{average} respectively. Note that we do not correct for multiple comparisons; such a correction would further reduce the significance of the results. We implement the comparison test for smoothed ROC curves detailed in~\citet{robin_proc_2011}.\begin{itemize}
        \item 1 month forecasts: \textit{Z} = 0.64, 1.09, 0.42, 0.67; \textit{p} = 0.26, 0.14, 0.34, 0.25
        \item 6 months forecasts: \textit{Z} = 0.41, 0.08, 0.70, 0.69; \textit{p} = 0.34, 0.47, 0.24, 0.25
    \end{itemize}
    \item The 95\% confidence intervals for the 1 month models are: \begin{itemize}
        \item \textit{escalation}: 0.66-0.95
        \item \textit{quad}: 0.63-0.95
        \item \textit{goldstein}: 0.62-0.93
        \item \textit{cameo}: 0.65-0.95
        \item \textit{average}: 0.65-0.95
    \end{itemize}
    \item The 95\% confidence intervals for the 6 month models are: \begin{itemize}
        \item \textit{escalation}: 0.64-0.93
        \item \textit{quad}: 0.60-0.90
        \item \textit{goldstein}: 0.68-0.93
        \item \textit{cameo}: 0.58-0.92
        \item \textit{average}: 0.60-0.92
    \end{itemize}
\end{itemize}

While a small p-value is used to reject the null hypothesis (in this case --- that the out-of-sample performance does not differ between the models being compared), a singular focus on a test for statistical significance at a pre-defined threshold can be harmful (see, for example \citet{imbens_statistical_2021}).
Blair and Sambanis do report performance evaluations for a variety of different model specifications. However, the purpose of such robustness checks is to determine whether model performance sensitive to the parameter choices; it is unclear whether it helps deal with issues arising from sampling variance. 
At any rate, Blair and Sambanis's results turn out to be highly sensitive to another modeling choice: the fact that they compute the AUC metric on the smoothed ROC curve instead of the empirical curve that their model produces. Smoothing refers to a transformation of the ROC curve to make the predicted probabilities for the war and peace instances normally distributed instead of using the empirical ROC curve (see \citet{robin_proc_2011}). This issue was pointed out by \citet{beger_reassessing_2021} and completely changes their original results; \citet{blair_is_2021} discuss it in their rebuttal.

\subsection{Overview of papers in \Cref{tab:significance}}
\label{sec:significance}
\Cref{tab:significance} provides the list of 12 papers included in our review, showing information about whether they report confidence intervals, conduct tests of statistical significance when comparing classifier performance, which metrics they report, the number of rows and the number of positive instances (i.e. instances of war/conflict) in the test set, and whether their main claim relies on out-of-sample evaluation of classifier performance. We detail information about the numbers we report in \Cref{tab:significance} below.
\begin{itemize}
    \item \textbf{\citet{hegre_forecasting_2016}}: We report the number of rows and number of positive instances of civil war incidence for the dates between 2001 and 2013 in the UCDP dataset, i.e. all years for which out-of-sample estimates are provided. We report the out-of-sample AUC performance difference for the Major conflict setting. Out-of-sample evaluation results are not included in the main text of the paper, hence we report that the paper's main claim does not rely on out-of-sample evaluations.
    \item \textbf{\citet{muchlinski_comparing_2016}}: We report the number of rows and number of positive instances of civil war onset for the dates after 2000 in the out-of-sample dataset provided by Muchlinski et al. We report the out-of-sample AUC performance difference between the Random Forests and the best Logistic Regression setting. Out-of-sample evaluation results are used to justify the performance improvement of using Random Forests models, hence we report that the paper's main claim relies on out-of-sample evaluations.
    \item \textbf{\citet{chiba_shape_2017}}: We report the total number of instances and the number of positive instances of governmental onsets in the years 2013-14 (the test set dates). We report the difference between the territorial onset AUC's reported in the paper. Note that while \citet{chiba_shape_2017} do report small number of data points that are used in one of their settings, they do not address how to estimate variance or perform tests of statistical significance. Out-of-sample evaluation results are not used as the main evidence of better performance in the main text of the paper, hence we report that the paper's main claim does not rely on out-of-sample evaluations.
    \item \textbf{\citet{colaresi_robot_2017}}:  We report the number of rows and onsets of civil war after the year 1988 (the test set dates). We report the out-of-sample AUC difference between the two random forests models compared in the paper. Out-of-sample evaluation results are used to justify the performance improvement of using an iterative method for model improvement, hence we report that the paper's main claim relies on out-of-sample evaluations.
    \item \textbf{\citet{hirose_can_2017}}: We report the number of locations included in the out-of-sample results. Since the paper does not attempt binary classification, we do not report the number of positive instances in this case. We report the out-of-sample performance gain of adding relative ISAF support to the baseline model in the IED attack setting of the paper. Out-of-sample evaluation results are used as important evidence of better model performance in the main text of the paper, hence we report that the paper's main claim relies on out-of-sample evaluations.
    \item \textbf{\citet{schutte_regions_2017}}: We report the number of rows in the entire dataset, since the paper uses k-fold cross validation and therefore all instances are used for testing. Since the paper does not attempt binary classification, we do not report the number of positive instances in this case. We report the out-of-sample normalized MAE difference between the population model and the best performing model compared in the paper. Out-of-sample evaluation results are used as important evidence of better model performance in the main text of the paper, hence we report that the paper's main claim relies on out-of-sample evaluations.
    \item \textbf{\citet{hegre_evaluating_2019}}: We report the number of rows and number of positive instances of civil war incidence for the dates between 2001 and 2013 in the UCDP dataset, i.e. all years for which out-of-sample estimates are provided. We report the out-of-sample AUC performance difference for the Major conflict setting. Out-of-sample evaluation results are not used as the primary evidence of better model performance in the main text of the paper, hence we report that the paper's main claim does not rely on out-of-sample evaluations.
    \item \textbf{\citet{hegre_views_2019}}: We report the number instances with state based conflict in the ViEWS Monthly Outcomes at PRIO-Grid Level data between 2015 and 2017 — the years for which the out-of-sample results are reported in the paper. We report the out-of-sample AUC performance difference for the state-based conflict setting. Out-of-sample evaluation results are used as the primary evidence of better model performance in the main text of the paper, hence we report that the paper's main claim relies on out-of-sample evaluations.
    \item \textbf{\citet{kaufman_improving_2019}}: We report the total number of rows and all instances of civil war incidence in the dataset used by Kaufman et al., since they use k-fold cross validation and therefore all instances are used for testing. We report the out-of-sample accuracy difference between the Adaboost and Logistic Regression settings. Out-of-sample evaluation results are used as the primary evidence of better model performance in the main text of the paper, hence we report that the paper's main claim relies on out-of-sample evaluations.
    \item \textbf{\citet{wang_comparing_2019}}: We report the total number of rows and onsets of civil war used in the dataset used by Wang since they use k-fold cross validation and therefore all instances are used for testing. We report the out-of-sample AUC performance difference between the Adaboost and Logistic Regression models. Out-of-sample evaluation results are used as the primary evidence of better model performance in the main text of the paper, hence we report that the paper's main claim relies on out-of-sample evaluations.
    \item \textbf{\citet{blair_forecasting_2020}}: We report the number of rows and onsets of civil war after the year 2007 (the test set dates). We report the out-of-sample AUC performance difference between the escalation and cameo models for the one-month base setting. Out-of-sample evaluation results are used as the primary evidence of better model performance in the main text of the paper, hence we report that the paper's main claim relies on out-of-sample evaluations.
    \item \textbf{\citet{hegre_can_2021}}: We report the number of rows and number of positive instances for civil war onset the dates between 2001 and 2018, i.e. all years for which out-of-sample estimates are provided. We don't report the out-of-sample performance difference because the paper does not perform comparisons between models. Out-of-sample evaluation results are used as the primary evidence of model performance in the main text of the paper, hence we report that the paper's main claim relies on out-of-sample evaluations.
\end{itemize}

\begin{table*}[t]
  \centering
  \begin{adjustbox}{width=\columnwidth,center} 
  \begin{tabular}{llllllll}
    \toprule
    Paper & CI? & {\begin{tabular}[x]{@{}l@{}}Stat. sig\\test?\end{tabular}} & Metric(s) & {\begin{tabular}[x]{@{}l@{}}Num. rows \\in test set\end{tabular}} & {\begin{tabular}[x]{@{}l@{}}Num. positive\\test set instances\end{tabular}} & {\begin{tabular}[x]{@{}l@{}}Main Claim\\OOS?\end{tabular}} & {\begin{tabular}[x]{@{}l@{}}OOS performance\\
    delta\end{tabular}}
     \\
    \midrule
    \citet{hegre_forecasting_2016} & No & No & AUC, Brier score & 2197 & 321 & No & 0.006
    \\
    \citet{muchlinski_comparing_2016} & No & No & AUC, F1 score & 896 & 19 & Yes & 0.04
    \\
    \citet{chiba_shape_2017} & No & No & AUC, Brier score & 4176 & 15 & No & 0.03
    \\
    \citet{colaresi_robot_2017} & No & No & AUC, Precision, Recall & 1778 & 29 & Yes & 0.02
    \\
    \citet{hirose_can_2017} & No & * & MAE, RMSE & 14,606 & --- & Yes & 0.16
    \\
    \citet{schutte_regions_2017} & No & No & MAE & 3744 & --- & Yes & 0.09
    \\
    \citet{hegre_evaluating_2019} & No & No & AUC & 2197 & 321 & No & 0.02
    \\
    \citet{hegre_views_2019} & No & No & {\begin{tabular}[x]{@{}l@{}}AUC, Brier score, AUPR, Accuracy,\\F1 score, cost-based threshold\end{tabular}} & 384,372 & 1848 & Yes & 0.01
    \\
    \citet{kaufman_improving_2019} & No & No & Accuracy & 6610 & 918 & Yes & 0.03
    \\
    \citet{wang_comparing_2019} & Yes & No & AUC, Precision, Recall & 6363 & 116 & Yes & 0.12
    \\
    \citet{blair_forecasting_2020} & No & No & AUC, Precision, Recall & 15,744 & 11 & Yes & 0.03
    \\
    \citet{hegre_can_2021} & Yes & No & AUC, AUPR, TPR/FPR & 3042 & 79 & Yes & ---
    \\
    \bottomrule
  \end{tabular}
  \end{adjustbox}
    \caption{A list of papers for which code and dataset were available, showing information about whether they report confidence intervals, conduct tests of statistical significance when comparing classifier performance, which metrics they report, the number of rows and the number of positive instances (i.e. instances of war or conflict or onset thereof) in the test set, and whether their main claim relies on out-of-sample evaluation of classifier performance. AUC = Area Under ROC, MAE = Mean Absolute Error, RMSE = Root Mean Squared Error, AUPR = Area Under Precision-Recall Curve, TPR = True Positive Rate, FPR = False Positive Rate, OOS performance delta = the performance difference for the most salient performance comparison reported in the paper (details in Section \ref{sec:significance}). *Hirose et al. state that the out-of-sample performance is significantly better in the Supplement of their paper, but we could not find the figure they cite as evidence of this claim in their Supplement.}
    \label{tab:significance}
\end{table*}

\section{Model info sheets for detecting and preventing leakage in ML-based science} 
\label{app:model_cards}
We include the model info sheet template as a Microsoft Word document on our website (\url{https://reproducible.cs.princeton.edu}). Here, we detail how model info sheets would address each type of leakage that we found in our survey, as well as the types of leakage we found in our case study of civil war prediction. 

\begin{itemize}
    \item \textbf{L1.1 No test set.} Model info sheets require an explanation of how the train and test set is split during all steps in the modeling process (Q9-17 of model info sheets).
    \item \textbf{L1.2 Pre-processing on training and test set.} Details of how the train and test set are separated during the preprocessing selection step need to be included in the model info sheet (Q12-13). This would address leakage due to incorrect imputation \citet{muchlinski_comparing_2016, wang_comparing_2019, colaresi_robot_2017}.
    \item \textbf{L1.3 Feature selection on training and test set.} Details of how the train and test set are separated during the feature selection step need to be included in the model info sheet (Q14-15).
    \item \textbf{L1.4 Duplicates in datasets.} Model info sheets require details of whether there are duplicates in the dataset, and if so, how they are handled (Q10).
    \item \textbf{L2 Model uses features that are not legitimate.} For each feature used in the model, researchers need to argue why the feature is legitimate to be used for the modeling task at hand (Q21). This addresses the leakage due to the use of proxy variables in \citet{kaufman_improving_2019}.
    \item \textbf{L3.1 Temporal leakage.} In case the claim is about predicting future outcomes of interest based on ML methods, researchers need to provide an explanation for why the time windows used in the training and test set are separate, and why data in the test set is always a later timestamp compared to the data in the training set (Q20). This addresses the temporal leakage in \citet{kaufman_improving_2019, wang_comparing_2019}.
    \item \textbf{L3.2 Dependencies in training and test data.} Researchers need to reason about the dependencies that may exist in their dataset and outline how dependencies across training and test sets are addressed (Q11).
    \item \textbf{L3.3 Sampling bias in test distribution.} Researchers need to reason about the presence of selection bias in their dataset and outline how the rows included for data analysis were selected, and how the test set matches the distribution about which the scientific claims are made (Q18-19).
\end{itemize}

\end{document}